\documentclass[lettersize,journal,compsoc]{IEEEtran}
\usepackage{bm}
\usepackage{amsmath,amsfonts}
\usepackage{amssymb}
\usepackage{graphicx}
\usepackage{algorithmic}
\usepackage{paralist}
\usepackage{array}
\usepackage{algorithm}
\usepackage[caption=false,font=normalsize,labelfont=sf,textfont=sf]{subfig}
\usepackage{textcomp}
\usepackage{stfloats}
\usepackage{url}
\usepackage{verbatim}
\usepackage{cite}
\usepackage{booktabs}
\usepackage{tabularx}
\usepackage{makecell}
\usepackage[numbers]{natbib}
\usepackage{relsize}
\usepackage{rotating}
\usepackage{multirow}
\usepackage{diagbox}
\usepackage{stackengine}
\usepackage{balance}
\usepackage{natbib}
\usepackage{caption}
\usepackage{threeparttable}
\usepackage{tikz}
\usepackage{nicematrix}
\usetikzlibrary{arrows,positioning,shapes.multipart} 
\tikzset{
	>=stealth',
	punkt/.style={
		rectangle,
		rounded corners,
		draw=black, thick,
		text centered,
		font=\small,
	},
	rect/.style={
		rectangle,
		draw=black, thick,
		text centered,
		font=\small,
	},
	pil/.style={
		->,
		thick,
		shorten <=2pt,
		shorten >=2pt,
		font=\small,
	},
}
\usepackage{tkz-graph}
\usepackage{xcolor,colortbl}
\newcommand{\LL}[2]{{\upshape[$_\mathsf{#2}$ }#1{\upshape]}}

\begin{document}

\title{Temporal Knowledge Graph Question Answering: A Survey}

\author{Miao Su, Zixuan Li, Zhongni Hou, Zhuo Chen, Long Bai, Xiaolong Jin, Jiafeng Guo, 
        \thanks{All authors are with the School of Computer Science and Technology, University of Chinese Academy of Sciences, Beijing 100049, China, and also with the CAS Key Lab of Network Data Science and Technology, Institute of Computing Technology, Chinese Academy of Sciences, Beijing 100864, China. 
        
        E-mail: \{sumiao22z, jinxiaolong\}@ict.ac.cn.}
\thanks{Manuscript received April 20, 2025;}}

\markboth{Journal of \LaTeX\ Class Files,~Vol.~14, No.~8, August~2021}%
{Shell \MakeLowercase{\textit{et al.}}: A Sample Article Using IEEEtran.cls for IEEE Journals}

\IEEEpubid{0000--0000/00\$00.00~\copyright~2021 IEEE}

\IEEEtitleabstractindextext{
\begin{abstract}
Knowledge Graph Question Answering (KGQA) has been a long-standing field to answer questions based on Knowledge Graphs (KGs). Recently, the evolving nature of knowledge has attracted a growing interest in Temporal Knowledge Graph Question Answering (TKGQA), an emerging task to answer temporal questions based on Temporal Knowledge Graphs (TKGs). However, this field grapples with ambiguities in defining temporal questions and lacks a systematic categorization of existing TKGQA methods. Motivated by this, this paper presents a thorough survey from two perspectives: the taxonomy of temporal questions and the categorization of TKGQA methodologies. Specifically, we first propose a detailed taxonomy of temporal questions explored in prior studies. Subsequently, we provide an in-depth review of TKGQA methodologies of two categories: semantic parsing-based and TKG embedding-based. 
We explicate advanced solutions as well as techniques used when answering temporal questions. Then, we discuss the potential impact of Large Language Models (LLMs) on TKGQA.
Building on the above review, we highlight potential research directions for TKGQA. This work aims to serve as a comprehensive reference for TKGQA and to inspire further research in this area.
\end{abstract}
\begin{IEEEkeywords}
Large Language Models, Temporal Knowledge Graph, Knowledge Base Question Answering, Knowledge Reasoning.
\end{IEEEkeywords}
}
\maketitle

\section{Introduction}
Knowledge Graphs (KGs) store enormous facts in the form of structured triples, i.e., \texttt{<subject, relation, object>}. Many real-world tasks, such as recommendation systems and question answering, have been brought about prosperity with the ability of commonsense understanding and reasoning of KGs~\cite{jiSurveyKnowledgeGraphs2022}.
Among them, Knowledge Graph Question Answering (KGQA) is a task that aims to answer natural language questions based on KGs~\cite{dong2015KGQA}. It has garnered significant attention from academia and industry due to its crucial role in various intelligent applications such as the Google search engine and Apple Siri~\cite{zhouCommonsenseKnowledgeAware2018}.


Previous knowledge graph research has predominantly focused on static facts that remain unchanged over time. However, temporal information is crucial, as structured knowledge is often valid only within specific periods. To address this limitation, Temporal Knowledge Graphs (TKGs) have been introduced, incorporating timestamps to represent the temporal dynamics of facts. As illustrated in Figure~\ref{fig:TKGQA-task}, the TKG associates each fact with a specific time point between 1997 and 2001 to denote when \textit{Harry Potter and the Philosopher’s Stone} received various awards. This temporal representation enables TKGs to capture the evolving nature of real-world events, facilitating more accurate reasoning and retrieval of time-sensitive information.

Based on TKG, Temporal Knowledge Graph Question Answering (TKGQA) aims to answer questions that involve temporal constraints or require timestamped answers, referred to as temporal questions in this paper. As illustrated in Figure~\ref{fig:TKGQA-task}, the example question contains the temporal constraint ``last'', which identifies the most recent event in a time sequence. To tackle TKGQA, models must capture the nuances of constraint and the chronological order of events, making it an increasingly prominent challenge in KGQA. However, TKGQA still faces two key challenges: 

\begin{figure}
    \centering
    \includegraphics[width=\linewidth]{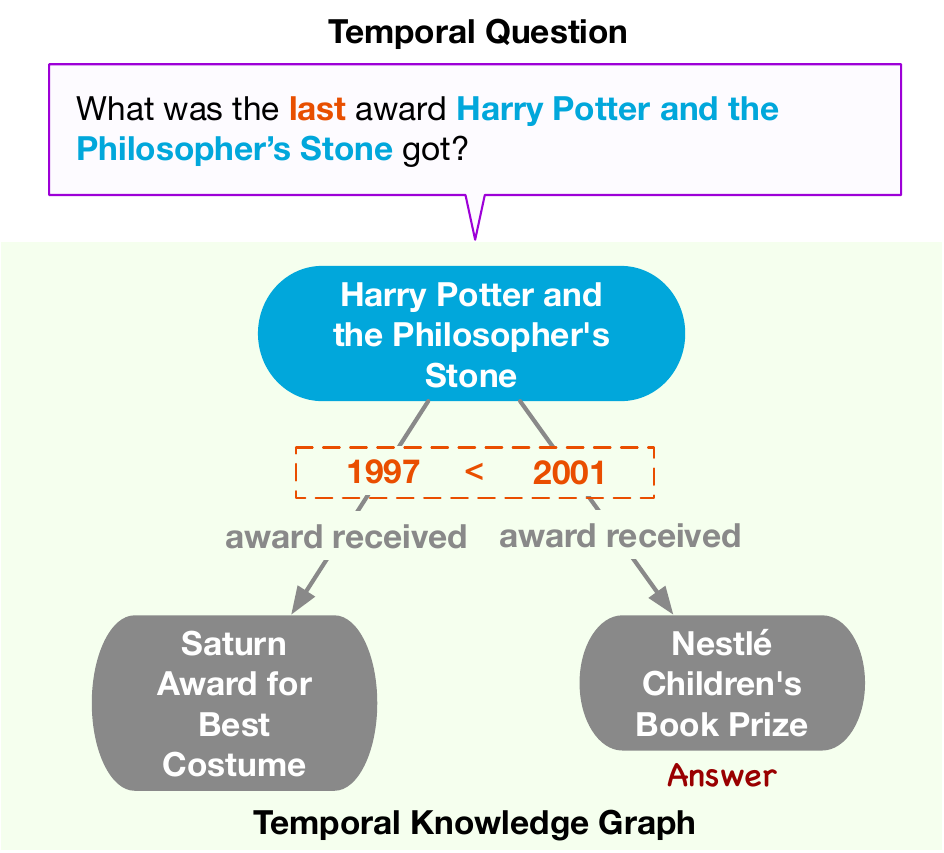}
    \caption{Given a temporal question (purple box), the TKGQA task aims to derive the answer from the underlying TKG (green box).}
    \label{fig:TKGQA-task}
\end{figure}

(1) \textbf{Inconsistency in the classification of temporal questions.} 
The existing research has not yet formed a unified framework for the type classification criteria of temporal questions. First, individual datasets often mix different classification criteria within a single framework. For example, as shown in Table \ref{tab:dataset}, CronQuestions applies the category \textit{SimpleTime} based on both the answer type and question complexity. At the same time, it also includes \textit{Before/After}, which is determined by content features within the same classification system. This cross-dimensional categorization leads to systematic ambiguity and inconsistency. Secondly, the Venn diagram of TempQuestions and CronQuestions shows that their classification systems have asymmetric overlaps, indicating a lack of logical compatibility between the existing classification systems.

Inconsistent classification of temporal questions can lead to several issues in model development and evaluation. If different works adopt conflicting classification schemes, it becomes difficult to compare methods fairly, hindering progress in the field.  This inconsistency can also lead to inefficiencies in model design—some models may apply inappropriate reasoning strategies, while others may lack necessary processing modules for certain question types. Consequently, the field urgently needs a well-defined and consistent classification framework for temporal questions.

\IEEEpubidadjcol

\begin{table}[htbp] 
    \caption{Question types and their Venn Diagram of TKGQA datasets.\label{tab:dataset}}
    \centering
    \begin{threeparttable}
    \resizebox{0.9\columnwidth}{!}{%
    \begin{tabular}{ccc}
    \toprule
        \textbf{Dataset} & \textbf{\makecell{Question\\Types}} & \textbf{Venn Diagram} \\ \midrule
        \makecell[c]{Temp\\Questions} & \makecell[c]{Explicit\\Implicit\\Ordinal\\Temp.Answer} & \makecell{\includegraphics[width=0.3\textwidth]{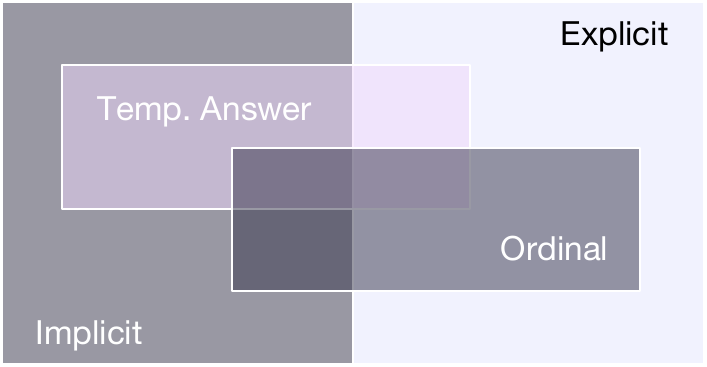}}\\ \midrule
        \makecell{Cron\\Questions} & \makecell{SimpleTime\\SimpleEntity\\Before/After\\First/Last\\TimeJoin} & \makecell{\includegraphics[width=0.3\textwidth]{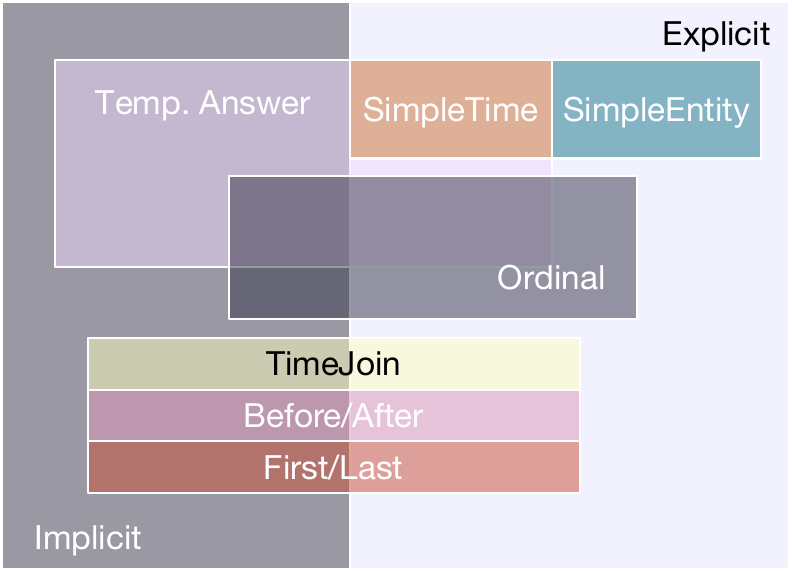}}\\ \midrule
        \makecell{MultiTQ} & \makecell{Equal\\Before/After\\First/Last\\Equal Multi\\Before Last\\After First} & \makecell{\includegraphics[width=0.2\textwidth]{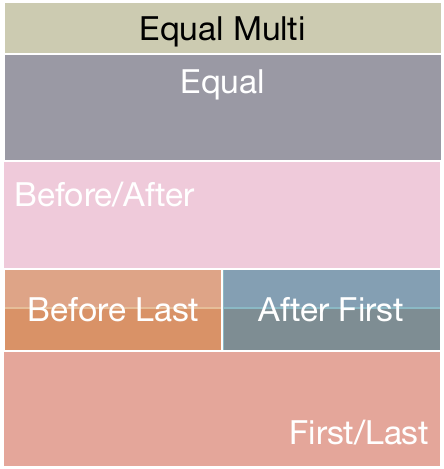}} \\ 
    \bottomrule
    \end{tabular}}
    \end{threeparttable}
\end{table}

(2) \textbf{Lack of systematic categorization of existing methods.}
 Existing surveys primarily focus on KGQA methods for static factual questions, adopting a two-tiered classification: \textbf{semantic parsing-based methods} that convert questions into formal logical queries (e.g., SPARQL), and \textbf{information retrieval-based methods} that match questions to answers via scoring mechanisms~\cite{fuSurveyComplexQuestion2020a,lanSurveyComplexKnowledge2021}. This framework effectively captures core technical distinctions in conventional KGQA where temporal dynamics are absent.

However, these classifications prove inadequate for TKGQA due to three inherent reasons:
\begin{inparaenum}[(i)]
\item The commonly used quadruple structure \texttt{<subject, relation, object, timestamp>} in TKGQA demands specialized temporal grounding mechanisms beyond static triple processing;
\item Diverse temporal question types require dedicated reasoning architectures;
\item The emergence of LLMs introduces new methodological paradigms that transcend traditional categorization.
\end{inparaenum}
Current surveys neither provide granular taxonomies for temporal-specific reasoning processes nor systematically analyze how LLM capabilities can be operated in temporal contexts~\cite{guKnowledgeBaseQuestion2022a}. This gap obstructs the mapping between question types and technical solutions, thereby limiting the ability to select and refine methods in a targeted manner.

To address the above challenges, this paper provides a thorough survey from two perspectives: a taxonomy of temporal questions and a dedicated categorization for TKGQA methods. Our main contributions can be summarized as follows:
\begin{itemize}
    \item We first establish a unified taxonomy that encompasses existing temporal question types and definitions, providing a standardized reference that could be widely adopted and further extended.
    
    \item We systematically categorize existing methods into semantic parsing-based, TKG embedding-based, and LLM-based. Furthermore, we explicate key procedures and techniques especially demanded by various temporal questions. This refined reclassification not only better aligns with the technical principles of TKGQA but also acknowledges emerging trends, such as LLMs’ potential to augment temporal reasoning. 
    
    \item We identify the temporal question types that each method can solve and summarize them in a table. This helps analyze the focus of existing methods and highlight question types that lack attention. Based on this review, we also explore future research directions.

    \item To the best of our knowledge, this is the first comprehensive survey on the TKGQA task. By serving as a comprehensive reference for TKGQA, this work aims to stimulate further research and foster innovation in the field. 
\end{itemize}

The rest of this survey is organized as follows. In §\ref{sec-pre}, we provide detailed definitions of TKGQA and relevant concepts. In §\ref{sec-question}, we classify temporal questions across all datasets based on 
question content (§\ref{question type}), answer type (§\ref{answertype}), and complexity (§\ref{complexity}).
In §\ref{sec-method}, we introduce the three main categories of TKGQA methods; in §\ref{sec-sp}, we detail semantic parsing-based methods, while in §\ref{sec-tkge}, we elaborate on TKG embedding-based methods; furthermore, we dive into LLM-based methods in §\ref{sec-llm}; in §\ref{taskcoverage}, we align each method with the specific types of questions it is designed to solve, providing a detailed summary table. We also introduce the evaluation metrics for TKGQA tasks (§\ref{sec-metrics}) and present a leaderboard illustrating the latest research progress (§\ref{leaderboard}).
In §\ref{future}, we explore new frontiers, summarizing their challenges and highlighting opportunities for further research. We conclude this survey in §\ref{conclusion}. Furthermore, in Appendix \ref{sec-appendix}, we provide a detailed description of the existing TKGQA datasets (§\ref{sec-datasets}), including their underlying temporal knowledge graphs. 

\section{Preliminary}\label{sec-pre}
This section introduces the key concepts and definitions related to the TKGQA task and its formulation.

\noindent\textbf{Temporal Knowledge Graph.}  Formally defined as $\mathcal{G} =  (\mathcal{E}, \mathcal{R}, \mathcal{T}, \mathcal{F})$, where $\mathcal{E}$, $\mathcal{R}$, $\mathcal{T}$, and $\mathcal{F}$ represent the sets of entities, relations, timestamps, and facts, respectively~\cite{caiSurveyTemporalKnowledge2024}. Each temporal fact \( f \in \mathcal{F} \) encodes relationships between entities with explicit timestamps, typically represented as $(s,r,o, \tau)$, quadruples where $s,o\in \mathcal{E}, r\in \mathcal{R},$ and $\tau\in\mathcal{T}$.

TKGs support multiple representation forms of temporal facts to capture temporal information effectively:
\begin{itemize}
\item \textbf{Compound Value Types (CVTs)}: In Freebase~\cite{bollackerFreebaseCollaborativelyCreated2008}, CVTs are attributes that consist of multiple related values. For example, the \texttt{marriage} predicate is a CVT with attributes like \texttt{marriage.spouse} and \texttt{marriage.date}.
\item \textbf{Timestamped Triples}: Triples with temporal annotations, e.g., \texttt{<Malia Obama, date\_of\_birth, 1998-07-04>}
\item \textbf{N-array Tuples}: Encode multi-attribute facts through qualifiers, e.g., \texttt{<Barack Obama, position held, President of the United States; start date, 20-01-2009; end date, 20-01-2017>}
\item \textbf{Quintuples}: Compact representation of interval-based facts, e.g., \texttt{<Barack Obama, position held, President of the United States, 2009, 2017>}. 
\item \textbf{Events}: Encode significant events such as \texttt{<WWII, significant event, occurred, 1939, 1945>}~\cite{saxenaQuestionAnsweringTemporal2021}.
\end{itemize}

Notably, some representations can be decomposed into standard quadruples. For instance, the presidential term quintuple \texttt{<Barack Obama, position held, President of the United States, 2009, 2017>} can be decomposed into annual quadruples: \texttt{<Barack Obama, position held, President of the United States, 2009>}, \texttt{<…, 2010>}, …, \texttt{<…, 2017>}. Such normalization of time-based facts ensures that the system can be applied across various TKGs.

\noindent\textbf{Temporal Question.}
 A temporal question contains at least one temporal constraint or requires timestamps as its answer~\cite{jiaTempQuestionsBenchmarkTemporal2018}. 
 
The \textit{temporal constraint}, serving as the key distinguishing text span in temporal questions, specifies a condition tied to a particular time point or interval that the answer and its supporting evidence must satisfy. This constraint typically combines a \textit{temporal expression} (e.g., 2023) and a \textit{temporal signal} (e.g., after) that co-occur as contiguous spans within the question text.
 
To dissect this structure further, \textit{temporal expressions} refer to time points or intervals in temporal questions, which can vary in granularity (e.g., May 11th, 2024)~\cite{pustejovskyTimeMLRobustSpecification,huangDomainSensitiveTemporalTagging2018}. Complementing temporal expressions, \textit{temporal signals} indicate the relationships between them and act as trigger words, imposing constraints on the answer or supporting evidence (e.g., in, after, or during).

\noindent\textbf{Temporal Knowledge Graph Question Answering.} Given a TKG $\mathcal{G}$ and a temporal question $q$ expressed in natural language, the task of TKGQA aims to infer the correct answer from the structured data. The answers consist of entity sets $\left\{e \mid e \in \mathcal{E}\right\}$ or timestamps $\left\{\tau \mid \tau \in \mathcal{T}\right\}$ within $\mathcal{G}$.

\noindent\textbf{TKGQA Datasets.} A TKGQA dataset $\mathcal{D}$ comprises pairs of questions and answers $(q, a)$. Different datasets may include additional information, such as entity identifiers~\cite{neelamSYGMASystemGeneralizable2021}, temporal expressions within $q$, or question types. 

\begin{table}[]
\caption{Knowledge representation form of TKGQA datasets.\label{tab:knowledge}}
\begin{tabular}{lll}
\toprule
\textbf{Dataset}       & \textbf{Underlying TKG}        & \textbf{\makecell{Knowledge Representation \\ Form}}                                                   \\ \midrule
TempQuestions & Freebase   & CVTs                                                                            \\ \midrule
TimeQuestions & Wikidata   & \begin{tabular}[c]{@{}l@{}}Timestamped Triples and\\ N-array Tuple\end{tabular} \\ \midrule
CronQuestions & Wikidata   & Quintuples or Quadruples                                                        \\ \midrule
MultiTQ       & ICEWS05-15 & Quadruples \\ \bottomrule                
\end{tabular}
\end{table}

Since different datasets are constructed from different knowledge graphs, their representation form of temporal facts also varies. We list a few examples in Table~\ref {tab:knowledge}. 

\section{Taxonomy of Temporal Questions\label{sec-question}}

\begin{figure*}[h]
    \centering    \includegraphics[width=\textwidth]{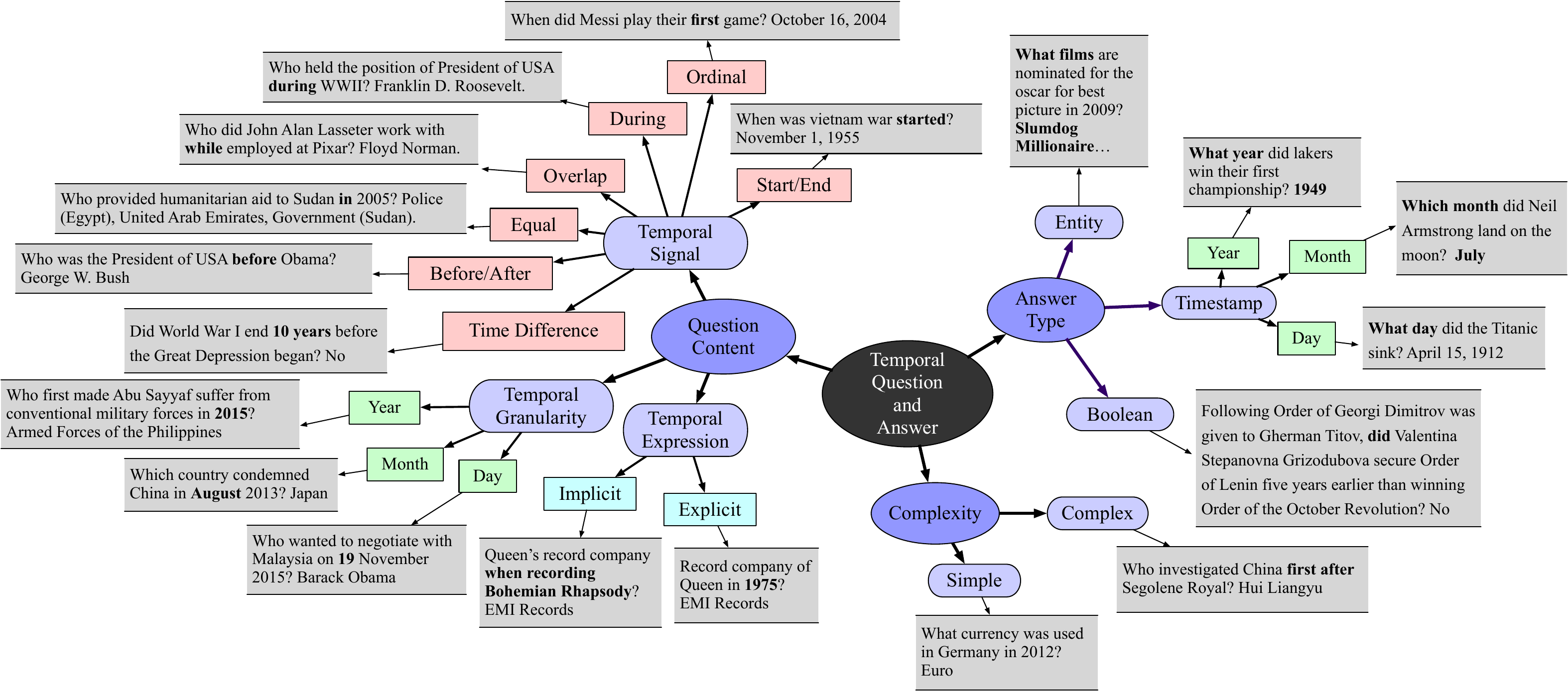}
    \caption{Taxonomy of temporal questions from three aspects, including (a) Question Content; (b) Answer Type, and (c) Complexity. Each grey box contains an example of a question-answer pair.}
    \label{fig:taxonomy}
\end{figure*}

We categorise temporal questions according to three key aspects:
\begin{inparaenum}[(i)]
\item Question Content: Questions are classified based on various time-related dimensions, which fundamentally shape how they should be addressed;
\item Answer Type: We organize questions by the nature of their expected answers. Unlike most existing KGQA research, which primarily targets entity-type answers, TKGQA must accommodate a wider range of answer types (e.g., timestamps), which we argue should be foundational to general KGQA systems;
\item Complexity: Following the convention in KGQA~\cite{huStatetransitionFrameworkAnswer2018, luoKnowledgeBaseQuestion2018}, we also categorize questions based on their complexity.
\end{inparaenum}

The overall taxonomy of temporal questions is illustrated in Figure~\ref{fig:taxonomy}, and detailed introductions of each category are as follows.

\subsection{Question Content\label{question type}}
As shown in Figure~\ref{fig:taxonomy} (a), we classify the questions based on the time-related phrases present in the question.

 \emph{Temporal Granularity.} Questions can be categorized by the temporal granularity of their temporal expressions. The granularity \emph{year} is the most frequently used, followed by \emph{day} and \emph{month}. In cases where no explicit temporal granularity is provided, it typically aligns with the timestamp granularity of the underlying TKG or is inferred from context. Less common granularities, such as centuries, hours, or minutes, are not currently represented in existing TKGs; our focus remains on the granularities that are currently supported.

{\emph{Temporal Expression.}}
Questions can be classified as either \emph{explicit} or \emph{implicit} based on the nature of their temporal expressions.

All time points can be normalized to a standard format, such as 2024-08-09. An \textit{explicit temporal expression} can be normalized without additional context (e.g., September 2023 can be normalized to 2023-09). In contrast, an \textit{implicit temporal expression} requires additional knowledge to be converted into a specific time point or interval~\cite{jiaTempQuestionsBenchmarkTemporal2018}. For instance, an event name (e.g., the Paris Olympics) needs to be contextualized. It is crucial to specify and normalize the implicit temporal expression and align it with the TKG timestamp to answer a question correctly.

{\emph{Temporal Signal.}}
Temporal signals indicate the temporal relationships between temporal expressions.
We simplify Allen’s Interval Algebra for temporal reasoning~\cite{allenMaintainingKnowledgeTemporal1983} into six types of relations: \textit{Before/After}, \textit{Equal}, \textit{Overlap}, \textit{During}, \textit{Start/End}, and \textit{Ordinal}. These relations focus on the comparison of timestamps. Additionally, timestamps can be added or subtracted. Therefore, \textit{Time Difference} is introduced to constrain the time difference between facts~\cite{zhangMusTQTemporalKnowledge2024}. 

Figure~\ref{fig:taxonomy} (a) provides examples of each type of temporal signal, while a summary of the formula definitions for these temporal signal types is presented in Table~\ref{tab:constraint_type}.

\begin{table*}[htbp]
    \caption{Formula definition of signal types with explanations. \label{tab:constraint_type}}
    \centering
    \resizebox{\textwidth}{!}{%
    \begin{tabular}{c c l}
    \toprule
        \textbf{Signal Type} & \textbf{Formula Definition} & \textbf{Explanation} \\ 
        \midrule
        Before & $end_{ans}\leq begin_{cons}$ & The answer ends before the constraint begins. \\ 
        \addlinespace
        \midrule
        After & $begin_{ans}\geq end_{cons}$ & The answer starts after the constraint ends. \\ 
        \addlinespace
        \midrule
        Equal & $begin_{ans}=begin_{cons}, end_{ans}=end_{cons}$ & The answer and the constraint have identical start and end times. \\ 
        \addlinespace
        \midrule
        Overlap & \begin{tabular}{@{}c@{}}
     $begin_{ans}\leq end_{cons}\leq end_{ans}$ or \\
     $begin_{ans} \leq begin_{cons} \leq end_{ans}$
   \end{tabular} & Part of the answer overlaps with the constraint range. \\ 
        \addlinespace
        \midrule
        During & $begin_{cons}\leq begin_{ans} \leq end_{ans} \leq end_{cons}$ & The answer is completely within the constraint range. \\ 
        \addlinespace
        \midrule
        Start & $begin_{ans}=begin_{cons}\leq end_{ans} \leq end_{cons}$ & The answer starts at the same time as the constraint but may end earlier. \\ 
        \addlinespace
        \midrule
        End & $begin_{cons} \leq begin_{ans} \leq end_{cons} = end_{ans}$ & The answer ends at the same time as the constraint but may start later. \\ 
        \addlinespace
        \midrule
        Ordinal & $Rank(ans) = {Position}_{cons}$ & The answer is the specific position in a chronologically ordered list. \\
        \addlinespace
        \midrule
        Time Difference & $Point_{ans}=Point_{cons} + \Delta t$ & The exact time difference between the answer and the constraint is $\Delta t$. \\ 
        \bottomrule
    \end{tabular}
    }
\end{table*}

\subsection{Answer Type}
\label{answertype}
TKGQA datasets predominantly categorize answer types into two principal classes: \emph{entity sets} and \emph{timestamp sets}. The temporal granularity of timestamp answers (e.g., year, month, or day) varies according to contextual constraints inherent in either the question content or the TKG. 

Although this binary categorization suffices for conventional datasets, several specialized benchmarks have expanded answer taxonomies to address advanced reasoning requirements. For instance, ForecastTKGQuestions~\cite{dingForecastTKGQuestionsBenchmarkTemporal2022} diverges by incorporating both \textit{yes/no} responses and a modified binary classification framework (\textit{unknown}) derived from triple validation tasks. This dataset further integrates \textit{multiple-choice} answer structures (i.e., requiring selection from predefined options)  to facilitate evidence-based reasoning verification. Similarly, MusTQ introduces \textit{boolean} answers for temporal difference verification and \textit{numeric} responses to quantify time interval reasoning. These innovations broaden the scope of evaluative criteria, enabling more nuanced assessments of complex temporal reasoning capabilities.

\subsection{Complexity\label{complexity}}
KGQA defines question complexity such that single-hop questions are considered simple, while multi-hop and constrained questions are classified as complex~\cite{baoConstraintBasedQuestionAnswering2016}. Within this framework, all temporal questions inherently fall under the category of complex questions due to their reliance on temporal constraints. However, this broad classification is shared with other complex question types, such as comparative and quantitative queries. Temporal questions present unique challenges stemming from temporal granularity alignment, temporal expression implicitness, temporal signals understanding, and multiple-constraint composition. This distinct combination of complexity drivers necessitates dedicated treatment when defining and evaluating temporal question answering capabilities~\cite{fuSurveyComplexQuestion2020a,lanSurveyComplexKnowledge2021}. Therefore, we propose a two-tiered taxonomy for temporal questions to better address their specific characteristics.

\noindent{\emph{Simple temporal questions.}} 
Simple temporal questions involve direct queries with explicit temporal expressions, where answers are straightforwardly extracted from the TKG without the need for complex reasoning or comparisons.
For example, ``What currency was used in Germany in 2012?'' queries the fact \texttt{<Germany, currency, Euro, 2012>}. ``When did World War II start?'' retrieves the singular event timestamp \texttt{1939}.

\noindent{\emph{Complex temporal questions.}} 
Requiring multi-step reasoning about temporal relationships, these questions present four distinct challenges that demand specialized handling:

\begin{itemize}
    \item \textbf{Granularity Alignment:} Resolves mismatches between query scope and TKG timestamp precision.\\
    Example: For ``Which films did Hitchcock release in 1960?'', the system must recognize that \texttt{1960-10-07}  (release date of movie \textit{Psycho} in TKG) satisfies the annual timescale constraint.
    
    \item \textbf{Implicit-to-Explicit Conversion:} Maps implicit temporal expressions to explicit temporal points or intervals using TKG context.\\
    Example: ``Who was U.S. President immediately before Obama?'' requires deriving "before 2009" from TKG fact \texttt{<Barack\_Obama, President\_of, USA, 2009, 2017>}.

    \item \textbf{Temporal Signal Understanding:} Identifies and resolves temporal signals anchored to specific events.\\
    Example: ``Who succeeded Lyndon B. Johnson as U.S. President after John F. Kennedy's assassination in 1963?'' requires understanding \textit{after} in the context of a known historical event.
    
    \item \textbf{Multi-Constraint Composition:} Combines multiple constraints.\\
    Example: ``After the Danish Ministry's delegation visited Iraq in 2015, which European leader first visited the country?'' combines temporal ordering (\textit{after 2015}) and ordinal reasoning (\textit{first}).
\end{itemize}
\section{Categories of TKGQA Methods\label{sec-method}}
We categorize TKGQA methods into three primary paradigms: {\textbf{Semantic Parsing-based} \textbf{(SP-based)}, \textbf{TKG Embedding-based} \textbf{(TKGE-based)} methods, and \textbf{Large Language Model-based (LLM-based)} approaches. SP-based and TKGE-based methods currently dominate the field, with LLM-based techniques forming an emerging research direction.

SP-based methods operate by parsing natural language questions into structured, machine-interpretable representations (i.e., temporal queries), which are subsequently executed against the TKG to derive answers. These approaches emphasize explicit symbolic reasoning through formal query construction.

TKGE-based methods, conversely, adopt an embedding-driven paradigm: they first project both the question and candidate answers into latent vector spaces, then employ scoring functions to rank candidates based on their semantic and temporal compatibility. This paradigm implicitly models temporal dependencies through learned embeddings.

A fundamental distinction between SP-based and TKGE-based methods lies in how they derive answers. SP-based approaches rely on symbolic query execution for direct answer retrieval from the graph structure, whereas TKGE-based methods employ probabilistic scoring of latent representations to identify plausible answers. This dichotomy reflects broader tensions between explicit symbolic reasoning and implicit neural inference in knowledge-aware NLP systems.

LLM-based methods represent an emerging but transformative direction, capitalizing on the emergent reasoning capabilities and linguistic generalizability of large language models~\cite{xiong2024largelanguagemodelslearn}. These methods offer a complementary strength to SP-based and TKGE-based approaches. 

In the following subsections (Sec~\ref{sec-sp},~\ref{sec-tkge} and~\ref{sec-llm}), we will provide details on these categorisations.

\subsection{Semantic Parsing-based Methods\label{sec-sp}}
SP-based methods aim to convert natural language questions into TKG queries for answer retrieval. 
As depicted in Figure~\ref{fig:sp-2}, SP-based methods typically encompass three primary stages in the TKGQA task: question parsing, TKG grounding, and query execution and refinement. 

The \textit{question parsing} stage first performs semantic and syntactic analysis on the input temporal question, parsing it into an ungrounded query. This ungrounded query captures the key semantic information and temporal constraints of the question. Compared to the original natural language question, it typically presents a more structured representation. 
\textit{TKG grounding} then aligns the components of the ungrounded query with the elements in the TKG. After grounding, the ungrounded query is translated into executable queries, which may include entities, relations, and timestamps from the TKG.
Finally, the answer is obtained via \textit{query execution and refinement}. 

Subsequent subsections provide a systematic exposition of each phase and introduce the unique challenges and solutions faced by the existing methods in each stage. 

\begin{figure}[ht]
    \centering    \includegraphics[width=\columnwidth]{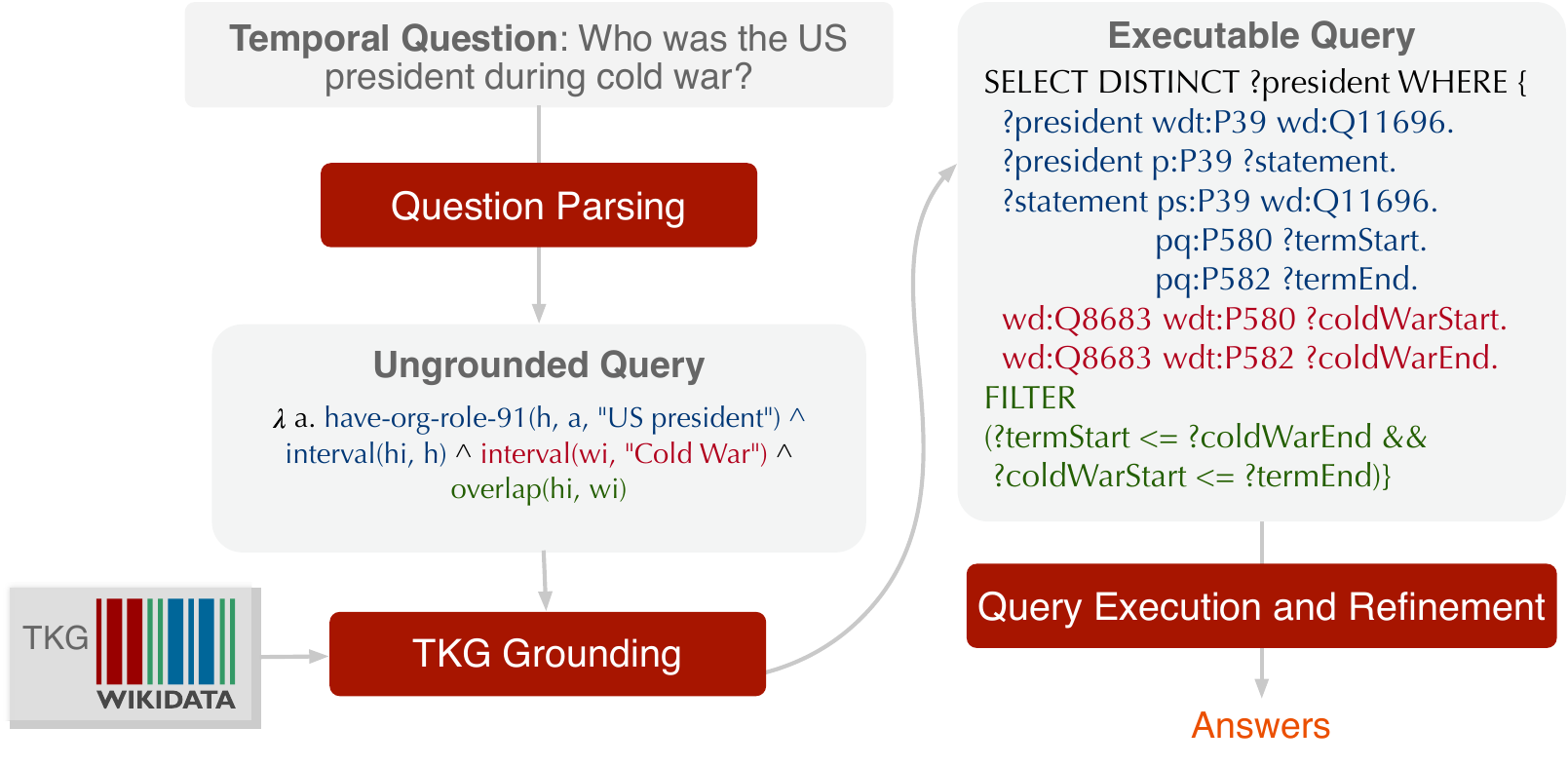}
    \caption{Overall procedure of SP-based methods. The representation forms of ungrounded query and executable query are $\lambda$-calculus and SPARQL, respectively.}
    \label{fig:sp-2}
\end{figure}

\subsubsection{Question Parsing}
Question parsing converts natural language questions into ungrounded queries, capturing semantic, syntactic, and temporal constraints for reasoning. These ungrounded queries may adopt varying representation forms across systems. Here, ``representation form'' denotes a structured format or expressive language that conveys the meaning of the questions. The ungrounded query serves as a ``sketch'' of the final executable query, abstracting away TKG-specific details like concrete entities, relations, or fully instantiated temporal constraints.

Before parsing, researchers must first select or design an appropriate representation form, a step we refer to as \textit{representation form selection and design}. Based on the chosen form, researchers then design the corresponding parsing model. We classify question-parsing methods into three categories: \textit{logical form-based parsing}, \textit{template-based parsing}, and \textit{AMR-based parsing}. These methods are discussed in detail in the following section.

\textbf{Representation Form Selection and Design.} A well-designed representation form facilitates accurate question interpretation and ensures compatibility with downstream grounding and reasoning.

In semantic-parsing methods for KGQA, there are already many logical forms that can be used to represent queries~\cite{kamathSurveySemanticParsing2019}. These include graph query languages (e.g., SPARQL~\cite{borrotoSPARQLQAv2SystemKnowledge2023}),  logic-oriented approaches (e.g., $\lambda$-calculus derivatives~\cite{caiLargescaleSemanticParsing2013}) and programming languages (e.g., S-expressions~\cite{guThreeLevelsGeneralization2021}, Prolog, or KoPL). Existing KGQA methods develop well-designed parsers for these logical forms. One simple way is to adopt these KGQA parsers to the TKGQA task. We call this line of methods \textit{logical form-based parsing}.

Apart from the representations mentioned above, Abstract Meaning Representation (AMR)~\cite{banarescuAbstractMeaningRepresentation2013} is a semantic representation language focusing on representing the meaning of natural language. It models sentence semantics as rooted directed acyclic graphs, where nodes represent concepts (e.g., entities, events) and edges denote semantic relationships. AMR explicitly encodes temporal constraints via the ``:time'' relation and temporal signals (e.g., before). It can also standardise temporal expressions through canonical forms like date-entity (e.g., 1960) and date-interval (e.g., the 20th century). These features make it suitable for representing temporal information. Since AMR tools have been developed for years and reached high results~\cite{bevilacquaRecentTrendsWord2021}, AMR parsers usually serve as the initial parsing tool to reduce the need for end-to-end training data~\cite{kapanipathiLeveragingAbstractMeaning2020}. We summarise these methods as \textit{AMR-based Parsing}.

Although the above representations can achieve complex operations, they lack specifically designed time operators required for time reasoning.
Other studies aim to build new representations that explicitly and concisely model temporal structures for TKGQA tasks while completely expressing the semantics of the question~\cite{chen2024self}. One direct way is to design templates with elements that need to be extracted from the question.
Ding~\cite{dingSemanticFrameworkBased2023} proposes the Semantic Framework of Temporal Constraints (SF-TCons), which systematically analyses the possible interpretation of temporal constraints in a way similar to a query graph. For the question “\textit{Which movie did Alfred Hitchcock \LL{direct}{Event_1} \LL{in}{Signal_1} \LL{1960}{Time_1}?}”, SF-TCons generates the graph template in Figure~\ref{fig:logical_form}, specifying that the temporal scope of the “direct” event must include 1960 (as inferred from the temporal signal “in”). This comparison structure is identified as 
    \begin{align*}
\textsc{Compare}&\langle\textsc{Includes},\mathrm{time}(\textit{``direct''}),\textit{``1960''}\rangle
    \end{align*}
where the comparative word (i.e., include) and timestamps are extracted and filled according to the question.
FAITH~\cite{jiaFaithfulTemporalQuestion2024} introduces a Time-aware Structured Frame (TSF) with two slot categories: general QA components (question entity, relation, answer type) and temporal QA components (temporal signal, category, value) to model temporal content explicitly. This template-like representation form captures the core components or elements in the question. 
These representations can be viewed as templates that need to be filled with phrases in the questions. They can be parsed via \textit{template filling}.

Extending existing logical forms is also one way to support temporal reasoning:
Prog-TQA extends the Knowledge-oriented Programming Language (KoPL)~\cite{caoKQAProDataset2022} with temporal operators like \texttt{FilterByTimePoint(Facts, "2012")}, enabling concise temporal query implementation compared to SPARQL-based approaches. SYGMA adopts Typed $\lambda$-Calculus~\cite{zettlemoyer2012learning}, which supports the addition of new higher-order functions. Apart from default functions, they add a few other functions such as interval, overlap, before, after, and so on, to support temporal reasoning. 

\textbf{Logical Form-based Parsing.}
To use the existing semantic parser in KGQA methods, TEQUILA~\cite{jiaTEQUILATemporalQuestion2018} decomposes temporal questions into non-temporal sub-questions (yielding candidate answers) and temporal sub-questions (yielding temporal constraints), using temporal signal words as decomposition pivots. These sub-questions are then reformulated using four predefined templates to improve readability. It leverages KGQA engines AQQU~\cite{AQQU} and QUINT~\cite{QUINT} to parse the sub-questions into SPARQL queries and execute them on the KG.

\textbf{Template-based Parsing.}
Template-based parsing methods use predefined templates as representation forms, which are filled with specific question phrases to parse the input question.  
Considering the temporal nature of TKG and temporal questions, existing methods focus on filling time-anchored information except for entities and relations commonly concerned in KGQA.

SF-TQA~\cite{dingSemanticFrameworkBased2023} fine-tunes a BERT model to annotate temporal elements within questions. Temporal signals serve as triggers for specific SF-TCons, which define corresponding temporal constraints and interpretation structures. Some temporal elements are identified based on predefined temporal relations, such as those specified by TimeML, while others are selected based on their proximity to the detected temporal signals (e.g., the nearest temporal expression).
When the question involves a comparison structure defined by SF-TCons (e.g., equal), the system normalises the TimeML relation types into standard algebraic expressions (such as =) to facilitate precise temporal reasoning.
FAITH also has a fine-tuned Seq2Seq model (BART) to populate general QA slots in TSF. Training data derives from: (i) distant supervision (question entities/relations), (ii) KB-type lookups (answer types), and (iii) benchmark annotations (temporal signals/categories). Explicit temporal values are extracted via SUTime~\cite{changSUTIMELibraryRecognizing} and regex matching. At the same time, implicit constraints require resolution through a recursive mechanism: an InstructGPT-generated~\cite{ouyangTrainingLanguageModels2022a} intermediate question (e.g., converting “Queen’s lead singer after Freddie Mercury?” to “When was Freddie Mercury lead singer of Queen?”) is fed back into the temporal QA system for explicit value extraction. 

In general, when filling in the representation form of templates, the time expressions shown in the question can be annotated using regular expressions, SUTime, and other simple tools. Existing tools such as TimeML or information in the dataset can be used for temporal signals. Existing methods can also fine-tune a language model with lightweight parameters to implement the relevant functionality. For implicit temporal constraints, further processing is needed using parsing methods such as question rewrite.

\begin{figure}[t]
    \centering
    \includegraphics[width=0.7\columnwidth]
    {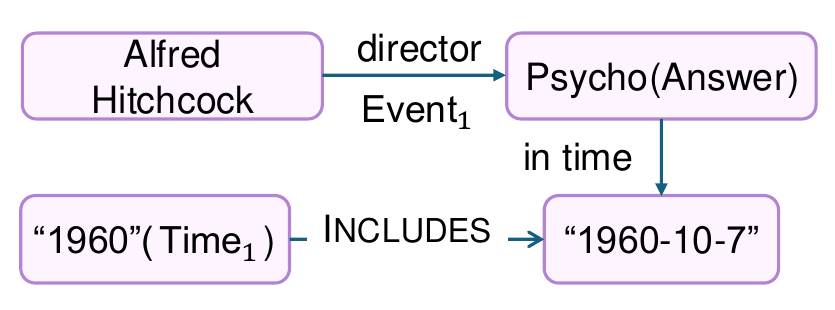}    
    \caption {Semantic framework of temporal constraints for question ``\textit{Which movie did Alfred Hitchcock \LL{direct}{Event_1} \LL{in}{Signal_1} \LL{1960}{Time_1}?}''.}
    \label{fig:logical_form}
  \end{figure}

\textbf{AMR-based Parsing.}
AMR provides a rich semantic representation of natural language by modelling sentence semantics. Its long-standing development and high performance make AMR parsers an effective initial tool during multi-stage parsing. 

Frameworks like SYGMA~\cite{neelamSYGMASystemGeneralizable2021}, AE-TQ~\cite{longComplexQuestionAnswering2022}, and Kannen~\cite{kannenBestBothWorlds2023} leverage pre-trained or self-trained AMR parsers for initial question parsing. However, raw AMR can not support temporal reasoning. To address this, these works refine AMR through mappings to task-specific representations:
SYGMA applies heuristic rules to convert the AMR frame into the corresponding typed $\lambda$-expression, a logic-oriented representation focusing on knowledge representation and reasoning. To support temporal reasoning, they add temporal functions such as ``interval'', ``overlap'', ``before'', ``after'', and so on. For instance, AMR’s ``:time'' relation, specifically the “nested-frame” structure, is mapped to the “overlap” function in the $\lambda$-expression.
To facilitate direct querying of facts from the TKG, AE-TQ designs Semantic Information Structures (SISs) containing {subject, relation, object, timestamp, and temporal signal word}. It uses rule-based mappings of AMR to SIS tuples. For example, AMR's ``:ord'' relation (e.g., first) maps to an SIS’s temporal signal slot, enabling structured temporal reasoning over TKGs. 


\subsubsection{TKG Grounding}
Following initial question parsing, the system generates an ungrounded query, a high-level representation lacking concrete TKG elements. To convert it into an executable form, the system must ground its components by aligning phrases with specific entities, relations, and timestamps in the TKG.
When mismatches occur between the representation form of the ungrounded query and the required executable query language, a query translation step is performed to reconcile the differences and ensure compatibility with the TKG.

\textbf{Ground Key Elements.} 
Key elements in TKG include entities, relations, events, and timestamps, which must be accurately identified and linked to construct meaningful queries. 

For entity and relation linking, off-the-shelf Named Entity Linking (NEL) models and relation linking techniques are commonly employed~\cite{chenTemporalKnowledgeQuestion2023, neelamSYGMASystemGeneralizable2021, kannenBestBothWorlds2023}. Methods such as BERT-based representation similarity~\cite{longComplexQuestionAnswering2022}, fuzzy matching with candidate selection~\cite{chenSelfImprovementProgrammingTemporal2024}, and distant supervision training of seq2seq models~\cite{jia2024faithful} are also frequently utilised. SF-TQA believes that the event in question during initial parsing can be grounded with entities as well as relationships in TKG. To ground these events, they classify them into two categories: \textit{nominal}, which are directly linked to entities, and \textit{predicative}, which are associated with relations or their corresponding facts within the subgraphs of linked entities. This linking is implemented by scoring their serialisations with a BERT model.

Grounded timestamps can be directly extracted or normalized from the temporal expressions identified from natural questions.
For explicit timestamps, TEQUILA uses HeidelTime~\cite{strotgenHeidelTimeHighQuality2010} to tag TIMEX3 expressions in questions, while SF-TQA fine-tunes a BERT model to annotate temporal elements. Faith, on the other hand, employs SUTime~\cite{changSUTIMELibraryRecognizing} and regular expression matching to handle explicit temporal expressions in questions.

\textbf{Query Translation.} 
The ungrounded query generated during initial question parsing must be translated into a query language compatible with TKGs. SYGMA~\cite{neelamSYGMASystemGeneralizable2021} finds that different TKGs adopt various temporal relations; the ungrounded query needs to be translated into corresponding queries to be executed on different TKGs.
They address this by applying a set of temporal-specific translation rules. For example, it maps the “interval” predicate in $\lambda$-calculus to a SPARQL pattern that uses the Wikidata properties Start time (wdt:P580) and End time (wdt:P582) to represent temporal constraints. 

\subsubsection{Query Execution and Refinement}
After getting the executable query, query execution is conducted by an existing executor over the TKG. However, due to challenges such as suboptimal query generation, limitations in temporal logic parsing, and the inherent incompleteness of TKGs, the initial execution may fail to produce accurate or complete results.
To mitigate these issues, recent research has proposed several refinement strategies. \textit{Query refinement} focuses on improving the quality of the executable query itself. In parallel, \textit{TKG refinement} addresses the incompleteness of the knowledge graph. \textit{Execution result refinement} targets models that lack the capabilities of producing executable queries with correct temporal constraints, refining the raw execution outputs by explicitly applying temporal constraints to candidate answers. Together, these approaches improve the overall robustness and accuracy of temporal question answering over TKGs.

\textbf{Query Refinement.}
Formulating executable queries with precise temporal constraints over TKGs is challenging. Query refinement aims to improve the quality of generated queries, thereby enhancing the final reasoning performance.

To improve robustness against errors in executable queries, SF-TQA adopts a multi-query ranking framework. It generates multiple candidate queries for each question and uses a BERT-based scoring model to select the most appropriate one. Adversarial training is introduced to construct training examples for the scoring model. Negative samples include confusing queries (partial temporal answers with lower F1 scores) and irrelevant queries (no temporal overlap with the correct answers).
In contrast, Prog-TQA performs refinement through post-processing. It detects potential errors in KoPL programs, generates corrected versions, and identifies valid ones. These are then executed, and the resulting answers are aggregated using a bootstrapping strategy.

\textbf{TKG Refinement.}
TKG incompleteness refers to the absence of facts in the TKG, i.e., the temporal knowledge in TKG cannot cover all real-world facts. To mitigate this, Kannen~\cite{kannenBestBothWorlds2023} proposed a targeted temporal fact extraction technique to supplement knowledge from text for TKG refinement. Their method decomposes the original $\lambda$-expression into main-$\lambda$ and auxiliary-$\lambda$. The main-$\lambda$ represents the primary event in the question, while the auxiliary-$\lambda$ contains the temporal constraint. If the execution of the auxiliary-$\lambda$ fails due to missing facts, additional temporal information is retrieved using a reading comprehension question-answering model. This recovered information is then integrated into the reasoning process to complete the query and enhance overall performance.

\textbf{Execution Result Refinement.}
For models that do not natively support parsing complex temporal logic, temporal constraints cannot be directly enforced during query execution. Consequently, additional refinement of the execution results becomes necessary. This typically involves explicitly applying temporal constraints to the candidate answers obtained from query execution in order to filter and refine the final outputs.

TEQUILA first perform knowledge graph lookups to retrieve the temporal scopes of candidate answers. For instance, in Freebase, the temporal scope of the predicate \textit{marriage.spouse} is obtained by retrieving the corresponding \textit{marriage.date}. If the date is missing, TEQUILA selects alternative relevant temporal predicates using a similarity-based approach. It then conducts explicit temporal validation, systematically evaluating candidate answers based on constraint formulations summarised in Table~\ref{tab:constraint_type}.
AE-TQ adopts a similar two-step strategy. It first retrieves candidate quadruples using an earlier representation known as SIS, identifying associated facts and temporal constraints. Temporal constraints are then applied to filter and rank the candidates, from which the final answer is inferred.

\subsection{TKG Embedding-based Methods\label{sec-tkge}}
While SP-based methods excel at explicit temporal reasoning through predefined temporal logic, they face inherent limitations in handling complex temporal interactions and unseen event correlations. Complementary to these symbolic methods, TKG Embedding-based techniques adopt a neural paradigm that implicitly captures temporal-semantic regularities through continuous vector representations. 

As formalised in Figure~\ref{fig:tkge-2}, TKG Embedding (TKGE) based methods implement a multi-stage neural architecture comprising three coordinated components: TKG representation, question representation, and answer ranking. First, the TKG representation module embeds structural and temporal relationships within the knowledge graph into a continuous vector space. Concurrently, the question representation module transforms the input question into a dense embedding that captures its semantic intent and enhances it with graph knowledge. During the answer ranking phase, the question embedding is projected into $Q_{ent}$ and $Q_{time}$ for ranking entities and timestamps separately.

\begin{figure}[ht]
    \centering
    \includegraphics[width=0.85\columnwidth]
    {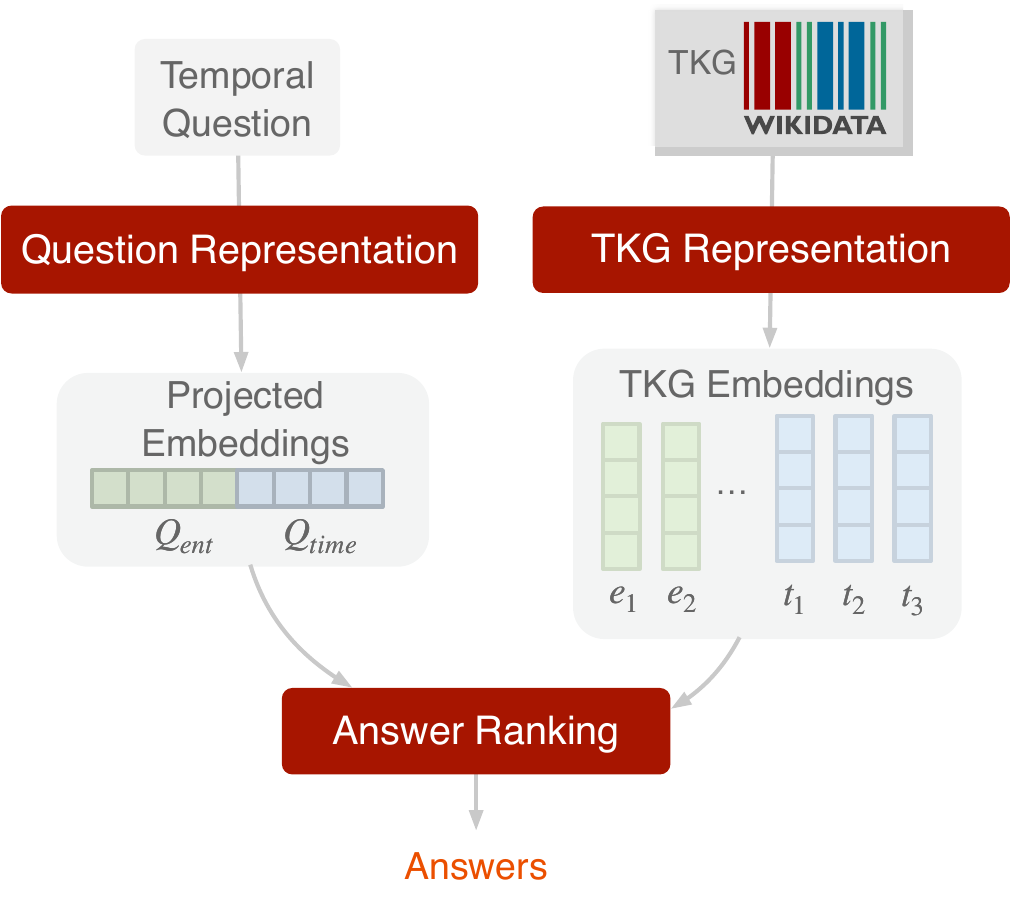}
    \caption{Overall procedure of TKGE-based methods.}
    \label{fig:tkge-2}
\end{figure}

\subsubsection{TKG Representation}
The TKG representation module seeks to capture the essence of TKG elements by transforming them into a low-dimensional vector space. Two predominant approaches for TKG representation are \textit{Graph Neural Network-based} and \textit{Decomposition-based} methods. Graph Neural Network-based TKG Representation utilises specialised Graph Neural Networks (GNNs) designed for graph-structured data, producing a variety of outputs depending on the task at hand~\cite{juComprehensiveSurveyDeep2024}. These networks are particularly effective in modeling complex relationships and dependencies within the graph, including temporal interactions.

In contrast, decomposition-based representation leverages tensor decomposition techniques to represent temporal knowledge as a four-dimensional tensor, where each dimension corresponds to the head
entity, relation, tail entity, and timestamp, respectively.
The embeddings generated for entities and timestamps are refined and enriched to form a pool of candidate answers, which can be used for downstream tasks.

\textbf{Graph Neural Network-based TKG Representation.}
Leveraging graph representation learning techniques such as Graph Convolutional Network (GCN) and Graph Neural Network (GNN), various methods have been developed to extract candidate answers from a question-specific subgraph. These approaches utilize message passing across nodes to propagate and aggregate information from neighbouring entities within the subgraph~\cite{sunOpenDomainQuestion2018a, yasunagaQAGNNReasoningLanguage2021}. The refined representations generated through this process are then employed to identify and select the most relevant answers.

The EXAQT framework~\cite{jiaComplexTemporalQuestion2021} uses a Relation Graph Convolutional Network (R-GCN) as the basic architecture, with innovations at three levels. In the subgraph construction stage, the Group Steiner Trees (GSTs) algorithm~\cite{liEfficientProgressiveGroup2016} is used to generate compact problem subgraphs, as each question keyword can match multiple nodes in the graph and naturally induces a terminal group. In the initialisation stage, it combines Wikipedia2Vec pre-trained embeddings~\cite{yamadaWikipedia2VecEfficientToolkit2020} and handcrafted temporal features, where the time-aware embeddings use the position encoding scheme from Zhang et al.~\cite{zhangTemporalContextAwareRepresentation2020} to map timestamps to a 64-dimensional vector space. In the temporal encoding integration aspect, EXAQT innovatively introduces a triple mechanism: a time attention module based on question type~\cite{jiaTEQUILATemporalQuestion2018}, a signal enhancement layer that incorporates temporal tags from news corpora~\cite{setzerTemporalInformationNewswire2001}, and a gated graph convolutional unit to control the propagation of temporal information.

\textbf{Decomposition-based TKG Representation.}
Common tensor decomposition techniques for TKG representation include Canonical Polyadic (CP) decomposition and Tucker decomposition. A summary of the most widely used decomposition-based methods for temporal knowledge graph embedding can be found in Appendix~\ref{TKGembedding}.

Motivated by the application of KG embeddings in KGQA, \textnormal{C{\smaller{RON}}KGQA}~\cite{saxenaQuestionAnsweringTemporal2021} initially encodes all elements of the TKG using the TComplEx model~\cite{lacroixTensorDecompositionsTemporal2020}, a decomposition-based TKGE method~\cite{caiTemporalKnowledgeGraph2023}. TComplEx is designed to complete queries such as \texttt{<US, has president, ?, 2012>}, making it very suitable for simple questions ``Who was the President of the United States in 2012?'' \textnormal{C{\smaller{RON}}KGQA} achieves very high accuracy on simple temporal questions requiring only one fact to answer, but falls short regarding questions requiring more complex reasoning. 
TSQA~\cite{shangImprovingTimeSensitivity2022} highlight that TComlEx focuses solely on the truth value of a quadruple, neglecting the order of different quadruples, which is essential for \textit{before/after} and \textit{ordinal} type temporal questions. Inspired by position embeddings in transformers~\cite{vaswaniAttentionAllYou2023}, they add position embeddings to timestamp embeddings and train TComplEx through an auxiliary task of determining temporal orders.
Chen et al.~\cite{chenMultigranularityTemporalQuestion2023} found that pre-trained TKG embeddings contain only semantic information at a single time granularity, which is inconsistent with real-world questions.
To address this inconsistency, they improve \textnormal{C{\smaller{RON}}KGQA} and propose MultiQA, which embeds multi-granularity temporal information. It concatenates day embeddings of trained TKGE models within each month or year interval, adds position vectors, and then fuses the information using the transformer for month and year embeddings.
MusTKGQA~\cite{zhangMusTQTemporalKnowledge2024} optimises the TKG representation in time and fact perspectives with sequence alignment and fact alignment, respectively. Sequence alignment enhances the learning of chronological order between timestamps, while fact alignment improves the awareness of the time boundary of interval facts.

\subsubsection{Question Representation}
\begin{figure}[ht]
    \centering
    \includegraphics[width=0.9\columnwidth]
    {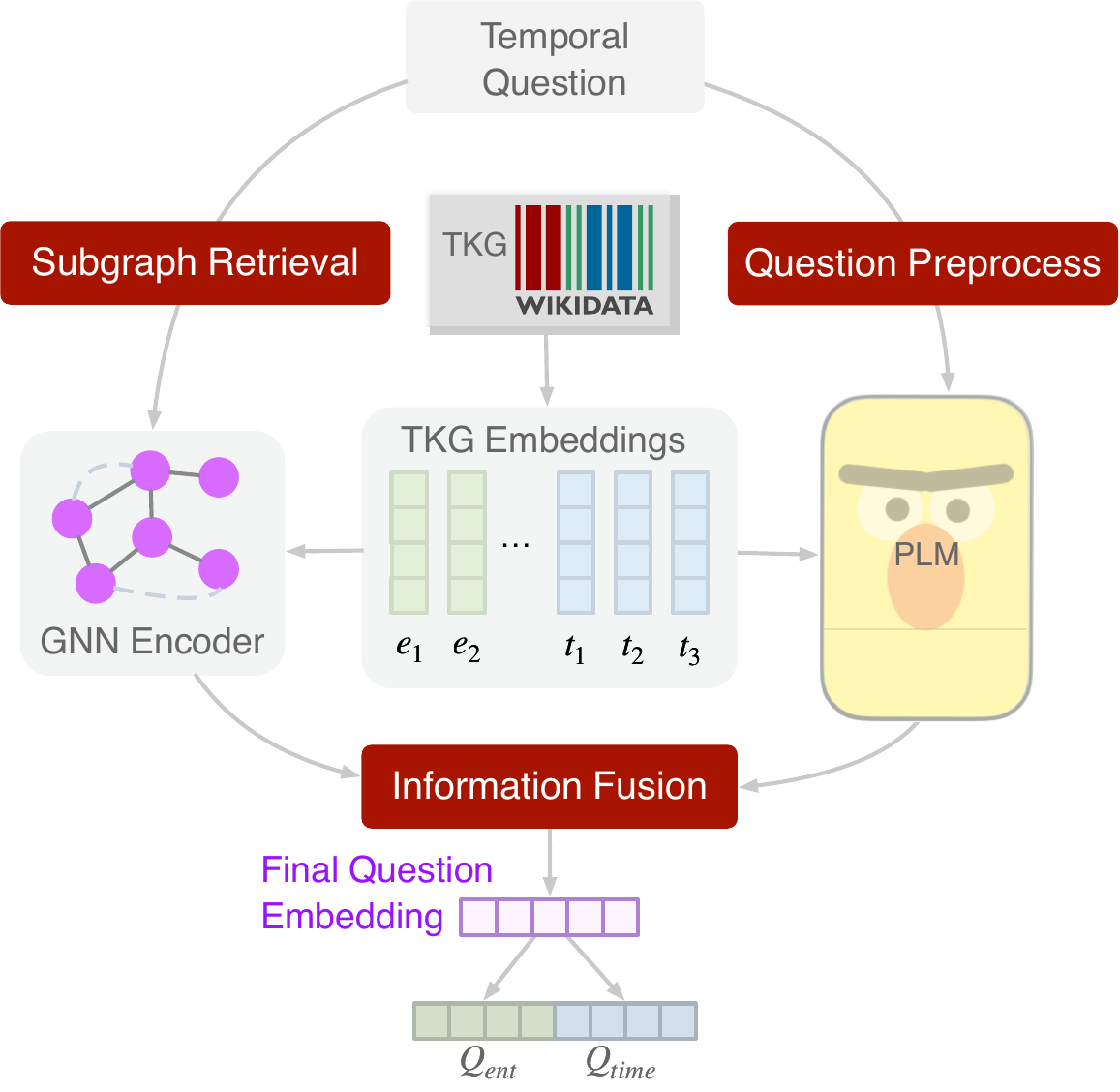}
    \caption{The overall question representation procedure of LGQA. The left is a process of graph structure integration; the right is a process of knowledge integration.}
    \label{fig:question-representation}
\end{figure}
The question representation module translates temporal questions into embeddings that capture semantics and time-sensitive information. 
As depicted in Figure~\ref{fig:question-representation}, PLMs~\cite{devlinBERTPretrainingDeep2019,liuRoBERTaRobustlyOptimized2019} usually offer a foundational question representation, which is augmented with TKG embeddings for \textit{knowledge integration}. Meanwhile, graph representation techniques encode a subgraph to incorporate structural information into PLM representations. Additionally, other methods~\cite{shangImprovingTimeSensitivity2022, ModelingTemporalSensitiveInformation2022,jiaoImprovingReasoningNetwork2023,duSemanticenhancedReasoningQuestion2024,chenTemporalKnowledgeGraph2022} enhance question understanding and representation through precise \textit{temporal constraint modelling}. 
These pieces of information are usually combined through information fusion to obtain the final question representation.

\textbf{Knowledge Integration.} 
Existing approaches employ two strategic layers to integrate structural and temporal knowledge into question embeddings: \textit{TKG Embedding fusion} and \textit{temporal knowledge enhancement}. The former incorporates outputs from TKG embedding models—typically entities and timestamps--into question semantics, while the latter directly integrates knowledge represented as quadruples.

\textit{TKG Embedding Fusion} employs a semantic grounding through PLM-TKG hybrid encoding. The process depicted on the right side of Figure~\ref{fig:question-representation} represents a simplified version of this approach. 
Saxena~\cite{saxenaQuestionAnsweringTemporal2021} pioneers this paradigm by generating BERT-based question representations ($\mathbf{Q}_B=\mathbf{W}_B \operatorname{BERT}\left(\mathbf{q}_0\right)$).
TempoQR~\cite{mavromatisTempoQRTemporalQuestion2022} advances this through entity-aware temporal grounding, replacing standard PLM token embeddings with TKG embeddings for entities/timestamps:

$$\mathbf{q}_{E_i}= \begin{cases} \mathbf{W}_E\mathbf{e}_\epsilon, & \text{if token }i\text{ is linked to an entity }\epsilon, \\ \mathbf{W}_E\mathbf{t}_\tau, & \text{if token }i\text{ is linked to a timestamp }\tau, \\ \mathbf{q}_{B_i}, & \text{otherwise.} \end{cases}$$

Furthermore, adding time scopes for each entity,  

$$\mathbf{q}_{T_i}= \begin{cases} \mathbf{q}_{E_i}, & \text{if token }i\text{ is not an entity,} \\ \mathbf{q}_{E_i}+\mathbf{t}_1+\mathbf{t}_2, & \text{if token }i\text{ is an entity,} \end{cases}$$
$t_1$ and $t_2$ implicitly define the time scope of the temporal constraint in the question, playing a crucial role in ranking candidate answers.
TempoQR first proposes to extract the question-specific temporal scope with two approaches. The ﬁrst retrieves the relevant information from the underlying TKG based on the annotated entities of the question. The second infers temporal information by solving a link prediction problem, obviating the need to access the TKG. Instead, TSIQA~\cite{xiaoModelingTemporalSensitiveInformation2022a} derives the time scope embedding of entities based on the assumption that entities with co-sharing relations correspond to related timestamps. Furthermore, MusTKGQA computes a unique time scope embedding for each question entity through an entity-time attention mechanism.

\textit{Temporal knowledge enhancement} adaptive fuses facts through multi-perspective attention. This mechanism allows the question embeddings to encode the relevant knowledge from TKGs with different degrees of importance, which can promote understanding.
To emphasize the importance of different knowledge for the question, JMFRN~\cite{huangJointMultiFactsReasoning2024} aggregates entity and timestamp information of retrieved facts using time-aware and entity-aware attention~\cite{vaswaniAttentionAllYou2023}.
TMA~\cite{liuTimeAwareMultiwayAdaptive2023} selects facts with similar semantics for three kinds of token-level attention. A gating mechanism integrates these representations to enhance the question embedding. $M^3$TQA~\cite{zhaM3TQAMultiViewMultiHop2024a} designs a multi-stage aggregation module, enabling asynchronous alignment and fusion of bidirectional heterogeneous information from the PLMs and GNNs.

\textbf{Graph Structure Integration.} This integration uses GNN-based models to further integrate the graphical structure into question embedding. As shown on the left side of Figure~\ref{fig:question-representation}, a question-specific subgraph is extracted to represent the question; the value of an edge in the graph is the concatenation of relation and timestamp, i.e., $r || t$, which is specific to TKGQA tasks. 

EXAQT enhances GRAFT-Net's R-GCN~\cite{sunOpenDomainQuestion2018} by concatenating temporal encodings of question categories and temporal signals with LSTM-based question representations. It is then updated together with entity embeddings through entity nodes. 
Temporal edge weighting emerges as a key innovation: TwiRGCN~\cite{sharmaTwiRGCNTemporallyWeighted2023} dynamically modulates RGCN message passing through question-dependent edge weights, enhancing messages through relevant edges and diminishing those from irrelevant ones. This intuitive scheme imposed soft temporal constraints on the messages passed during convolution and performs well on implicit and ordinal-type questions.
LGQA~\cite{liuLocalGlobalTemporal2023} implements dual-stram fusion—global PLM semantics interact with local graph patterns through transformer-based cross-attention. 
As shown in Figure~\ref {fig:question-representation},
the local information module uses the same multi-hop message-passing paradigm with TwiRGCN. 
The global information module encodes the question with PLM as TempoQR.
GenTKGQA~\cite{gaoTwostageGenerativeQuestion2024} leverages LLM’s extraction ability~\cite{sunChatGPTGoodSearch2023} to construct a question-specific subgraph and uses a pre-trained temporal GNN layer~\cite{velickovicGraphAttentionNetworks2018} to embed elements in the subgraph into ``virtual knowledge indicators'' to represent the question.

\textbf{Temporal Constraint Modeling.} Given the significance of temporal constraints, sophisticated techniques have been developed to explicitly model them, improving overall comprehension.

To enhance the model's sensitivity to temporal signal words, TSQA and TSIQA reverse them (e.g., replacing ``before'' with ``after'') to construct contrastive questions and apply both order loss and answer loss for contrastive learning.
Various approaches extract implicit temporal features from questions: CTRN~\cite{jiaoImprovingReasoningNetwork2023} uses multi-head self-attention, GCN~\cite{sunOpenDomainQuestion2018}, and CNN~\cite{potaBestPracticesConvolutional2020} to capture these features and fuse them with augmented BERT representations, while SERQA~\cite{duSemanticenhancedReasoningQuestion2024} integrates temporal constraint features computed from syntactic information in constituent and dependency trees~\cite{sunReorderThenParse2022, zhangFastAccurateNeural2020, wangModelingMultipleLatent2023, liangBiSynGATBiSyntaxAware2022} combined with Masked Self-Attention (MSA). 
To enhance the interpretability of reasoning on implicit temporal questions, SubGTR designs an implicit expression parsing module to rewrite their temporal constraints explicitly. 

\subsubsection{Answer Ranking}
The answer ranking module needs to determine the answer type and then rank temporal-related candidate answers based on the question and candidate answer representations.

\textbf{Determine Answer Type.}
To distinguish answer types, TwiRGCN introduces a gating mechanism that learns the likelihood that an answer is an entity or a timestamp given the question. JMFRN adds a type of discrimination loss based on the question representation.
Since the entities and timestamps obtained from the TKGE model are represented separately, most methods can directly score these two types of candidate answers without determining the answer type.

\textbf{Temporal Answer Scoring.}
Temporal answer scoring aims to establish a score function to rank the answer candidates based on previously obtained question representation and identified answer type.
The basic scoring function of TKDE methods (such as TComplEx)~\cite{saxenaQuestionAnsweringTemporal2021,mavromatisTempoQRTemporalQuestion2022,zhangMusTQTemporalKnowledge2024} establishes the preliminary correlation between the candidate answer and the question, and the subsequent enhancement techniques optimize it from different dimensions: such as applying temporal activation functions to satisfy  constraints~\cite{chenTemporalKnowledgeGraph2022}, introducing gating mechanisms~\cite{sharmaTwiRGCNTemporallyWeighted2023} and etc.

TSQA and TSIQA rank candidates simply based on their similarities to the entities or timestamps derived from the question.
C{\smaller{RON}}KGQA and TempoQR adapt TComplEx's quadruple scoring function $\phi(s,r,o,t)$, projecting question embeddings as dynamic relations. For a candidate entity $e$:
\begin{equation}
\label{eq:entity}
\phi_{\text{ent}}(e) = \mathfrak{R} \left( (\bm{u}_s, \bm{q}\bm{e}_{\text{ent}}, \bm{u}^\star_e, \bm{w}_t) \right).
\end{equation}
Temporal candidates $t\in T$ are scored similarly using time-projected embeddings.
CTRN enhances robustness through symmetric scoring with swapped subject-object pairs.
Except for TKGE-based scoring, SubGTR implements four temporal activation functions for different question types to better assess how well candidate answers satisfy the time constraints.
GenTKGQA fine-tunes LLMs to generate the most relevant answers through prompt-based inference.

\subsection{LLM-based Methods\label{sec-llm}}
Recent advances in LLMs have demonstrated remarkable comprehension and reasoning capabilities across diverse NLP tasks~\cite{achiam2023gpt}. This progress has spurred significant interest in harnessing LLMs for TKGQA. However, due to the relatively early-stage development of LLM research, investigations into their application for TKGQA remain scarce and constitute an emerging frontier. To bridge this gap, we systematically analyze the role of LLMs in both KGQA and TKGQA, proposing a taxonomy that roughly categorizes their applications into three paradigms: \textit{RAG Paradigm} (generate questions and answers in an RAG process), \textit{Semantic Parsing Paradigm} (parsing natural language into structured queries with constraints), and \textit{Agentic Reasoning Paradigm} (iteratively retrieving and reasoning as agents over KGs and TKGs). Our study evaluates LLM-based methodologies explicitly designed for TKGQA as well as those adaptable from KGQA frameworks, while also critically examining the untapped potential of LLMs in advancing TKGQA tasks.  

\subsubsection{RAG Paradigm}
Retrieval-augmented generation (RAG)~\cite{lewisRetrievalAugmentedGenerationKnowledgeIntensive2021}} combines pre-trained parametric and non-parametric memory for language generation. As shown in Figure~\ref{fig:LLMasGenerator}, the key idea is to retrieve relevant knowledge (through the knowledge retriever) from KGs and then fuse the retrieved knowledge (through prompt) into LLMs for answer generation.

Before retrieving facts, LLM can go through an optional question rewriting module that converts implicit questions into explicit ones to facilitate question understanding and knowledge retrieval. 
Qian et al.~\cite{qianTimeR4TimeawareRetrievalAugmented2024} observed that LLMs deduce correct answers more effectively in the answer generation phase with explicit questions as inputs.
Consequently, they proposed Time$R^4$, which rewrites the implicit questions by retrieving the necessary background facts, as illustrated in Figure~\ref{fig:LLMasGenerator}. 
This retrieval-rewrite strategy ensures that all questions contain explicit temporal constraints. 
LLMs can also utilise their inherent knowledge to generate questions involving common knowledge. For example, the question ``Who was the first president of the US after World War II?'' can be transformed into ``Who was the first president of the US after 1945?'', thereby addressing potential incompleteness in the TKG. 
Subsequently, they implement a retrieve-rerank module aimed at retrieving semantically and temporally relevant facts from the TKGs
and reranking them according to the temporal constraints. Finally, they fine-tune open-source LLMs by inputting the retrieved facts from TKGs and the generated explicit questions to generate the final answers.
Jia et al.~\cite{jia2024faithful} introduce FAITH, a Temporal QA system that operates across not only TKGs but also other diverse sources like text corpora and web tables. FAITH consists of three core stages: Temporal Question Understanding, Faithful Evidence Retrieval, and Explainable Heterogeneous Answering. In the question understanding stage, FAITH converts implicit constraints into explicit temporal values by generating intermediate questions and recursively invoking the FAITH system itself. For instance, to answer the question, “What was Queen’s record company when recording ‘Bohemian Rhapsody’?”, it is necessary to determine the time interval during which the recording took place (August 1975 – September 1975). This involves generating two intermediate questions: (i) ``What was the start date of Queen’s recording of `Bohemian Rhapsody`?'' and (ii) ``What was the end date of Queen’s recording of `Bohemian Rhapsody`?''. By answering these intermediate questions, the original query can be transformed into an explicit one: ``What was Queen’s record company during August 1975 – September 1975?''. This explicit question is further used for faithful evidence retrieval. Finally, FAITH harnesses explainable heterogeneous answering to deliver accurate answers accompanied by explanatory evidence.

During the generation process of the RAG paradigm. LLMs can function as answer generators, reasoning over retrieved facts to generate answers. GenTKGQA~\cite{gao2024two} identifies two main challenges in utilizing LLMs for the TKGQA task: question-relevant subgraph retrieval and complex-type question reasoning. To tackle the first challenge, GenTKGQA retrieves the relevant subgraph by employing query-related relation ranking and time mining. To address the second challenge, it introduces three innovative virtual knowledge indicators that integrate structural and temporal information with instruction learning. This approach enables LLMs to gain a deeper understanding of the graph structure and enhances their reasoning capabilities for complex temporal questions. Li et al.\cite{li2024large} observed that existing work often overlooks sentence-level semantic information embedded in TKGs. To address this issue, they extract relevant knowledge based on transformed sentences built from quadruples within TKGs using prompt learning. LLMs are employed to comprehend the query and its associated knowledge to generate the final answer.

 \begin{figure}
     \centering
     \includegraphics[width=0.8\linewidth]{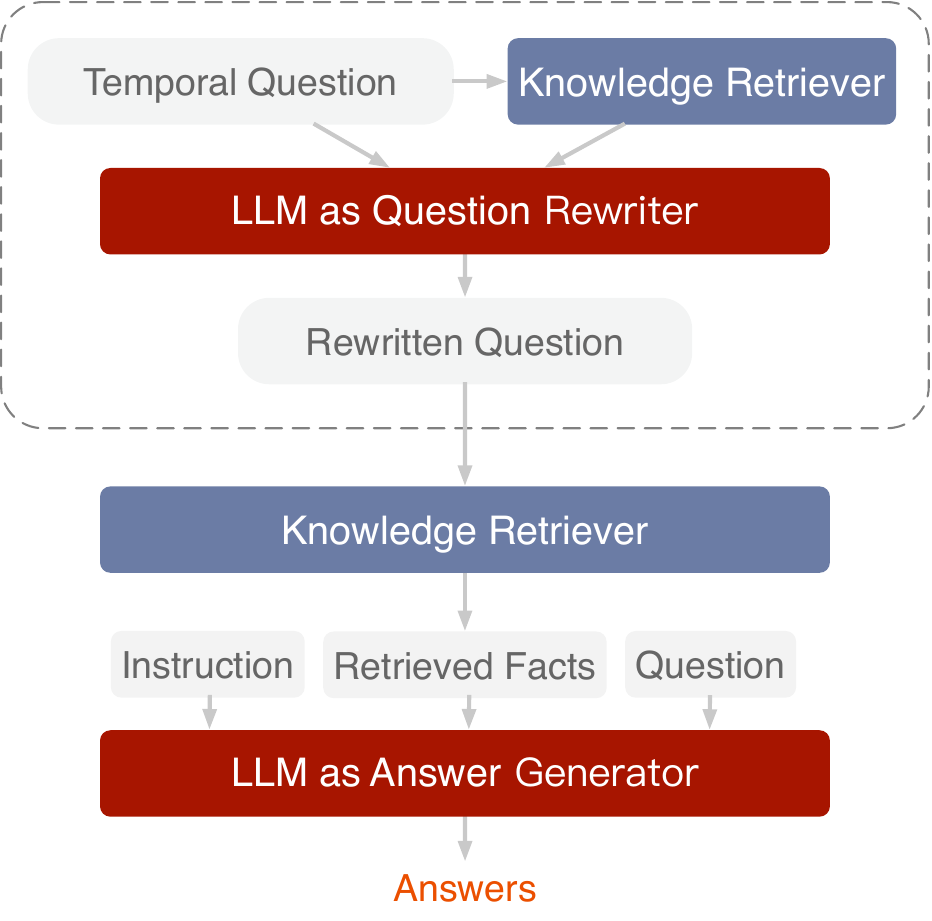}
     \caption{The general framework of adopting LLMs in retrieval augmented generation (RAG) paradigm for TKGQA task.}
     \label{fig:LLMasGenerator}
 \end{figure}

\subsubsection{Semantic Parsing Paradigm}
In the semantic parsing paradigm, LLMs can be used as a seq-to-seq model to transform the question into a structured query language. 

Motivated by SP-based methods that explicitly model constraints in questions by generating executable queries with symbolic operators, Chen et al.~\cite{chenSelfImprovementProgrammingTemporal2024} conduct a thorough analysis of time constraints within TKGQA. Subsequently, Chen et al. design corresponding temporal operators. Based on the designed operators, Chen et al. further propose a two-stage framework to parse the question into executable queries, namely Prog-TQA; it initially employs the in-context learning ability of LLMs to generate a draft for a given temporal question. Subsequently, the model links the draft with a specific TKG. 
Additionally, Prog-TQA incorporates an effective self-improvement strategy to enhance LLM’s comprehension of complex questions.
Many methods also follow this two-stage generation framework~\cite{liFewshotIncontextLearning2023,nieCodeStyleInContextLearning2024}, which aligns with the first two steps of the SP-based method mentioned in section~\ref{sec-sp}. 

The key challenge of LLMs as semantic parsers lies in their lack of structural proficiency in handling graph query languages. Since LLMs are pre-trained on large-scale natural language corpora rather than formal logic queries, KB-Coder~\cite{nieCodeStyleContextLearning2024} reframes the one-step generation of logical forms into a progressive, stepwise generation of Python function calls. This approach mitigates LLMs’ formatting errors and improves reliability. Similarly, Reasoning-Path-Editing (Readi)~\cite{chengCallMeWhen2024} bypasses direct logical form generation by first producing abstract reasoning paths and then grounding them on KGs. A reasoning path is a structured representation of a question’s logic—for example, the query “Which college did the daughter of Obama attend?” might map to the path ``[Obama] father\_of→college''. Experiments reveal that 46\%–60\% of LLM-generated reasoning paths can be successfully instantiated into executable KG queries. This method minimizes LLM inference calls and enhances computational efficiency.

\subsubsection{Agentic Reasoning Paradigm}
\begin{figure}
    \centering
    \includegraphics[width=\linewidth]{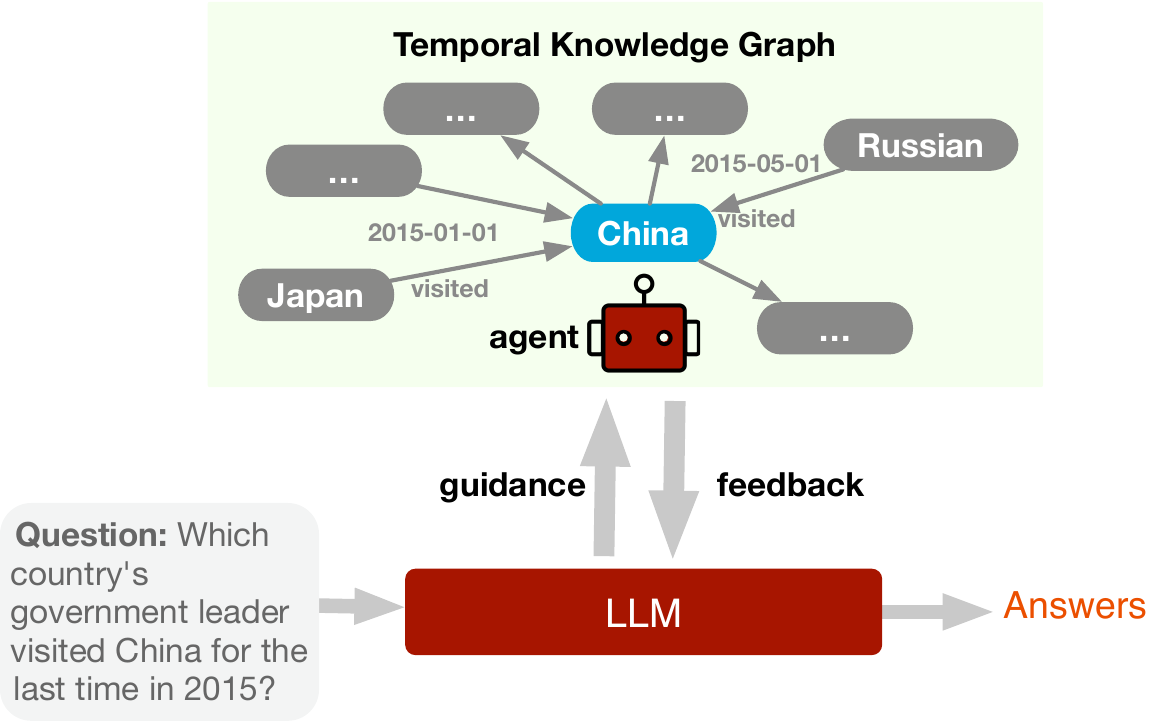}
    \caption{Using LLMs in agentic reasoning paradigm on TKG.}
    \label{fig:LLMAgent}
\end{figure}
Recent advancements have demonstrated that LLMs can be effectively adapted as autonomous reasoners (i.e., agents) for TKG interaction~\cite{chenTemporalKnowledgeQuestion2024}, we call this agentic reasoning paradigm. As illustrated in Figure~\ref{fig:LLMAgent}, the agent is capable of environmental perception, autonomous decision-making, and goal-directed actions~\cite{maesAgentsThatReduce1994,wooldridgeIntelligentAgentsTheory1995} in the agentic reasoning paradigm.

Generalized reasoning frameworks enable LLM-based agents to maintain robust performance on both unseen agent tasks and general language understanding benchmarks. They can be directly applied to TKGQA tasks. ReAct~\cite{yaoReActSynergizingReasoning2023} implements an interleaved generation mechanism for reasoning traces and task-specific actions. Reasoning traces help the model induce, track, and update action plans, while actions allow it to interface with and gather additional information from knowledge bases.
To enhance the agent abilities, AgentTuning~\cite{zengAgentTuningEnablingGeneralized2023} presents several specialized instruction tuning datasets to guide LLM agents to perform knowledge graph reasoning.

Most existing reasoning approaches based on LLMs primarily rely on entity-relationship chains as their foundation for logical inference. For example,
KSL~\cite{fengKnowledgeSolverTeaching2023} utilizes entity-centered path searches for answer generation but cannot handle temporal graphs due to its disregard for time. Similarly, StructGPT~\cite{jiangStructGPTGeneralFramework2023a} enables structural reasoning via API-mediated KG traversal but focuses solely on static triples. While Think-on-Graph~\cite{sunThinkonGraphDeepResponsible2024} offers a plug-and-play framework for iterative beam searches across KG entities, it also lacks explicit temporal reasoning mechanisms. However, these methods often neglect temporal constraints, limiting their ability to address temporal questions effectively. While such approaches are compatible with datasets built on static knowledge graphs (e.g., Freebase or Wikidata), their lack of temporal awareness hinders direct application to TKGQA. 

To bridge this gap, explicit temporal constraints could be integrated into the reasoning chains, enabling these frameworks to interpret temporal dependencies and adapt to temporal knowledge graphs. 
To equip LLMs with temporal logic reasoning ability, Chen et al.~\cite{chen2023temporal,valero2023comparing} first introduce an Abstract Reasoning Induction (ARI) framework. ARI divides temporal reasoning into two distinct phases: (1) a knowledge-agnostic phase for environmental interactions, allowing the LLM to make decisions based on abstract guidance of historical reasoning outcomes, and (2) a knowledge-based phase, where action choices are executed on specific TKGs to derive answers (figure~\ref{fig:LLMAgent} shows the guidance and feedback procedure).
These actions are customized for temporal knowledge graphs and temporal questions, including operations like getTime(head, rel, tail) and getBefore(entities, time). 
Additionally, ARI employs self-directed learning from historical reasoning outcomes, enabling LLMs to generalize methodologies (i.e., knowledge-agnostic instructions) across question types.

The integration of LLMs as agents for KG reasoning provides a flexible, training-free framework applicable to diverse LLMs and KGs, with the added benefit of interpretable reasoning processes. However, challenges persist in defining optimal action sets and policies for LLM agents. The synergy between LLMs and KGs remains an active research frontier, promising more sophisticated frameworks for temporal and multi-modal reasoning.

\begin{table*}
\caption{Comparison of Coverage across Question Categories in TKGQA Methods. \includegraphics[scale=0.2]{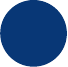} indicates that this method focuses on or specializes in solving this question category. \includegraphics[scale=0.2]{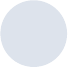} indicates this method can solve the corresponding question category. \label{align}} 
    \centering
    \scriptsize 
    \begin{NiceTabular}{|l|l|c|c|c|c|c|c|c|c|c|c|c|c|c|c|c|c|c|}
    \toprule[1 pt]
    \rule{0pt}{8pt}
    \Block{3-2}{\diagbox{\textbf{Method}}{\textbf{Category}}} & &\multicolumn{11}{c|}{Question Content} & \multicolumn{4}{c|}{Answer Type}&\multicolumn{2}{c|}{Complexity}  \\ \cline{3-19}
    
      &  &\multicolumn{3}{c|}{\shortstack{Time\\Granularity}}& \multicolumn{2}{c|}{\shortstack{Time\\Expression}}  & \multicolumn{6}{c|}{\shortstack{Temporal\\Signal}} & \Block{2-1}{Entity} &\multicolumn{3}{c|}{Time} & \Block{2-1}{Simple} & \Block{2-1}{Complex} \\ \cline{3-13}\cline{15-17}
      
     & &\begin{turn}{90}Year\end{turn}&\begin{turn}{90}Month\end{turn} & \begin{turn}{90}Day\end{turn}  & \begin{turn}{90}Explicit\end{turn} & \begin{turn}{90}Implicit\end{turn} &\begin{turn}{90}Overlap\end{turn}  & \begin{turn}{90}Before/After\end{turn} & \begin{turn}{90}Ordinal\end{turn} & 
          \begin{turn}{90}Equal\end{turn} &
          \begin{turn}{90}During/Include\end{turn} &
          \begin{turn}{90}Start/End\end{turn} &  & \begin{turn}{90}Year\end{turn} & \begin{turn}{90}Month\end{turn} & \begin{turn}{90}Day\end{turn} &  &  \\ \midrule[1pt]

        \multirow{5}{*}{\shortstack{Semantic\\ Parsing-\\based}} & TEQUILA~\cite{jiaTEQUILATemporalQuestion2018} & \includegraphics[scale=0.2]{figures/dots/1.pdf} & \includegraphics[scale=0.2]{figures/dots/1.pdf} & \includegraphics[scale=0.2]{figures/dots/1.pdf} & \includegraphics[scale=0.2]{figures/dots/1.pdf} & \includegraphics[scale=0.2]{figures/dots/1.pdf} & \includegraphics[scale=0.2]{figures/dots/1.pdf} & \includegraphics[scale=0.2]{figures/dots/1.pdf} & \includegraphics[scale=0.2]{figures/dots/1.pdf} & \includegraphics[scale=0.2]{figures/dots/1.pdf} & \includegraphics[scale=0.2]{figures/dots/1.pdf} & \includegraphics[scale=0.2]{figures/dots/1.pdf} & \includegraphics[scale=0.2]{figures/dots/1.pdf} & \includegraphics[scale=0.2]{figures/dots/1.pdf} & \includegraphics[scale=0.2]{figures/dots/1.pdf} & \includegraphics[scale=0.2]{figures/dots/1.pdf} & \includegraphics[scale=0.2]{figures/dots/1.pdf} & \includegraphics[scale=0.2]{figures/dots/1.pdf} \\ 
    
         & SYGMA~\cite{neelamSYGMASystemGeneralizable2021} & \includegraphics[scale=0.2]{figures/dots/1.pdf} & \includegraphics[scale=0.2]{figures/dots/1.pdf} & \includegraphics[scale=0.2]{figures/dots/1.pdf}  & \includegraphics[scale=0.2]{figures/dots/1.pdf} & \includegraphics[scale=0.2]{figures/dots/1.pdf} & \includegraphics[scale=0.2]{figures/dots/4.pdf} & \includegraphics[scale=0.2]{figures/dots/4.pdf} & \includegraphics[scale=0.2]{figures/dots/1.pdf} & \includegraphics[scale=0.2]{figures/dots/1.pdf} & \includegraphics[scale=0.2]{figures/dots/4.pdf} & \includegraphics[scale=0.2]{figures/dots/1.pdf} & \includegraphics[scale=0.2]{figures/dots/1.pdf} & \includegraphics[scale=0.2]{figures/dots/1.pdf} & \includegraphics[scale=0.2]{figures/dots/1.pdf} & \includegraphics[scale=0.2]{figures/dots/1.pdf} & \includegraphics[scale=0.2]{figures/dots/1.pdf} & \includegraphics[scale=0.2]{figures/dots/1.pdf} \\  
    
         & AE-TQ~\cite{longComplexQuestionAnswering2022}  & \includegraphics[scale=0.2]{figures/dots/1.pdf} &—&— & \includegraphics[scale=0.2]{figures/dots/1.pdf} & \includegraphics[scale=0.2]{figures/dots/1.pdf} & \includegraphics[scale=0.2]{figures/dots/4.pdf} & \includegraphics[scale=0.2]{figures/dots/4.pdf} & \includegraphics[scale=0.2]{figures/dots/4.pdf} & \includegraphics[scale=0.2]{figures/dots/1.pdf} & \includegraphics[scale=0.2]{figures/dots/1.pdf} & \includegraphics[scale=0.2]{figures/dots/1.pdf} & \includegraphics[scale=0.2]{figures/dots/1.pdf}  & \includegraphics[scale=0.2]{figures/dots/1.pdf} &—&—  & \includegraphics[scale=0.2]{figures/dots/1.pdf} & \includegraphics[scale=0.2]{figures/dots/4.pdf} \\  
    
         & SF-TQA~\cite{dingSemanticFrameworkBased2023}  & \includegraphics[scale=0.2]{figures/dots/1.pdf} & \includegraphics[scale=0.2]{figures/dots/1.pdf} & \includegraphics[scale=0.2]{figures/dots/1.pdf}  & \includegraphics[scale=0.2]{figures/dots/4.pdf} & \includegraphics[scale=0.2]{figures/dots/4.pdf} & \includegraphics[scale=0.2]{figures/dots/4.pdf} & \includegraphics[scale=0.2]{figures/dots/4.pdf} & \includegraphics[scale=0.2]{figures/dots/4.pdf} & \includegraphics[scale=0.2]{figures/dots/4.pdf} & \includegraphics[scale=0.2]{figures/dots/4.pdf} & \includegraphics[scale=0.2]{figures/dots/1.pdf} &  \includegraphics[scale=0.2]{figures/dots/1.pdf} & \includegraphics[scale=0.2]{figures/dots/1.pdf} & \includegraphics[scale=0.2]{figures/dots/1.pdf} & \includegraphics[scale=0.2]{figures/dots/1.pdf}  & \includegraphics[scale=0.2]{figures/dots/1.pdf} & \includegraphics[scale=0.2]{figures/dots/4.pdf} \\  
         
       & Best of Both~\cite{kannenBestBothWorlds2023} & \includegraphics[scale=0.2]{figures/dots/1.pdf} & \includegraphics[scale=0.2]{figures/dots/1.pdf} & \includegraphics[scale=0.2]{figures/dots/1.pdf} & \includegraphics[scale=0.2]{figures/dots/1.pdf} & \includegraphics[scale=0.2]{figures/dots/4.pdf} & \includegraphics[scale=0.2]{figures/dots/1.pdf} & \includegraphics[scale=0.2]{figures/dots/1.pdf} & \includegraphics[scale=0.2]{figures/dots/1.pdf} & \includegraphics[scale=0.2]{figures/dots/1.pdf} & \includegraphics[scale=0.2]{figures/dots/1.pdf} & \includegraphics[scale=0.2]{figures/dots/1.pdf} & \includegraphics[scale=0.2]{figures/dots/1.pdf} & \includegraphics[scale=0.2]{figures/dots/1.pdf} & \includegraphics[scale=0.2]{figures/dots/1.pdf} & \includegraphics[scale=0.2]{figures/dots/1.pdf} & \includegraphics[scale=0.2]{figures/dots/1.pdf} & \includegraphics[scale=0.2]{figures/dots/1.pdf} \\   \midrule[1 pt]  
        
        \multirow{16}{*}{\shortstack{TKG \\Embedding-\\based}} & CronKGQA~\cite{saxenaQuestionAnsweringTemporal2021}  & \includegraphics[scale=0.2]{figures/dots/1.pdf} &—&— & \includegraphics[scale=0.2]{figures/dots/1.pdf} & \includegraphics[scale=0.2]{figures/dots/1.pdf} & \includegraphics[scale=0.2]{figures/dots/1.pdf} & \includegraphics[scale=0.2]{figures/dots/1.pdf} & \includegraphics[scale=0.2]{figures/dots/1.pdf} &\includegraphics[scale=0.2]{figures/dots/1.pdf} & \includegraphics[scale=0.2]{figures/dots/1.pdf}&\includegraphics[scale=0.2]{figures/dots/1.pdf} & \includegraphics[scale=0.2]{figures/dots/1.pdf} & \includegraphics[scale=0.2]{figures/dots/1.pdf} &—&— & \includegraphics[scale=0.2]{figures/dots/4.pdf} & \includegraphics[scale=0.2]{figures/dots/1.pdf} \\ 
        
         & EXAQT~\cite{jiaComplexTemporalQuestion2021}  & \includegraphics[scale=0.2]{figures/dots/1.pdf} &—&—  & \includegraphics[scale=0.2]{figures/dots/4.pdf} & \includegraphics[scale=0.2]{figures/dots/1.pdf} & \includegraphics[scale=0.2]{figures/dots/1.pdf} & \includegraphics[scale=0.2]{figures/dots/1.pdf} & \includegraphics[scale=0.2]{figures/dots/1.pdf} & \includegraphics[scale=0.2]{figures/dots/1.pdf} & \includegraphics[scale=0.2]{figures/dots/1.pdf} & \includegraphics[scale=0.2]{figures/dots/1.pdf} & \includegraphics[scale=0.2]{figures/dots/1.pdf} & \includegraphics[scale=0.2]{figures/dots/4.pdf} &—&— & \includegraphics[scale=0.2]{figures/dots/1.pdf} & \includegraphics[scale=0.2]{figures/dots/1.pdf} \\ 
        
         & TempoQR~\cite{mavromatisTempoQRTemporalQuestion2022}  & \includegraphics[scale=0.2]{figures/dots/1.pdf} &—&—& \includegraphics[scale=0.2]{figures/dots/1.pdf} & \includegraphics[scale=0.2]{figures/dots/1.pdf} & \includegraphics[scale=0.2]{figures/dots/4.pdf} & \includegraphics[scale=0.2]{figures/dots/4.pdf} & \includegraphics[scale=0.2]{figures/dots/4.pdf} & \includegraphics[scale=0.2]{figures/dots/1.pdf} & \includegraphics[scale=0.2]{figures/dots/4.pdf} & \includegraphics[scale=0.2]{figures/dots/1.pdf} & \includegraphics[scale=0.2]{figures/dots/1.pdf}  & \includegraphics[scale=0.2]{figures/dots/1.pdf} &—&—  & \includegraphics[scale=0.2]{figures/dots/1.pdf} & \includegraphics[scale=0.2]{figures/dots/1.pdf} \\ 
        
         & TSQA~\cite{shangImprovingTimeSensitivity2022} & \includegraphics[scale=0.2]{figures/dots/1.pdf} &—&— & \includegraphics[scale=0.2]{figures/dots/1.pdf} & \includegraphics[scale=0.2]{figures/dots/1.pdf} & \includegraphics[scale=0.2]{figures/dots/1.pdf} & \includegraphics[scale=0.2]{figures/dots/4.pdf} & \includegraphics[scale=0.2]{figures/dots/4.pdf} & \includegraphics[scale=0.2]{figures/dots/4.pdf} &\includegraphics[scale=0.2]{figures/dots/4.pdf} & \includegraphics[scale=0.2]{figures/dots/1.pdf} & \includegraphics[scale=0.2]{figures/dots/1.pdf}  & \includegraphics[scale=0.2]{figures/dots/1.pdf} &—&—  & \includegraphics[scale=0.2]{figures/dots/1.pdf} & \includegraphics[scale=0.2]{figures/dots/4.pdf} \\ 
      
         & CTRN~\cite{jiaoImprovingReasoningNetwork2023}  & \includegraphics[scale=0.2]{figures/dots/1.pdf} &—&— & \includegraphics[scale=0.2]{figures/dots/1.pdf} & \includegraphics[scale=0.2]{figures/dots/1.pdf} & \includegraphics[scale=0.2]{figures/dots/4.pdf} & \includegraphics[scale=0.2]{figures/dots/4.pdf} & \includegraphics[scale=0.2]{figures/dots/4.pdf} & \includegraphics[scale=0.2]{figures/dots/1.pdf} & \includegraphics[scale=0.2]{figures/dots/4.pdf} & \includegraphics[scale=0.2]{figures/dots/1.pdf} & \includegraphics[scale=0.2]{figures/dots/1.pdf} & \includegraphics[scale=0.2]{figures/dots/1.pdf} &—&—  & \includegraphics[scale=0.2]{figures/dots/1.pdf} & \includegraphics[scale=0.2]{figures/dots/4.pdf} \\ 
        
         & SubGTR~\cite{chenTemporalKnowledgeGraph2022} & \includegraphics[scale=0.2]{figures/dots/1.pdf} &—&— & \includegraphics[scale=0.2]{figures/dots/1.pdf} & \includegraphics[scale=0.2]{figures/dots/4.pdf} & \includegraphics[scale=0.2]{figures/dots/4.pdf} & \includegraphics[scale=0.2]{figures/dots/4.pdf} & \includegraphics[scale=0.2]{figures/dots/4.pdf} & \includegraphics[scale=0.2]{figures/dots/1.pdf} & \includegraphics[scale=0.2]{figures/dots/4.pdf} & \includegraphics[scale=0.2]{figures/dots/1.pdf} & \includegraphics[scale=0.2]{figures/dots/1.pdf} & \includegraphics[scale=0.2]{figures/dots/1.pdf} &—&— & \includegraphics[scale=0.2]{figures/dots/1.pdf} & \includegraphics[scale=0.2]{figures/dots/4.pdf} \\ 
        
         & TwiRGCN~\cite{sharmaTwiRGCNTemporallyWeighted2023} & \includegraphics[scale=0.2]{figures/dots/1.pdf} &—&—& \includegraphics[scale=0.2]{figures/dots/1.pdf} & \includegraphics[scale=0.2]{figures/dots/4.pdf} & \includegraphics[scale=0.2]{figures/dots/1.pdf} & \includegraphics[scale=0.2]{figures/dots/1.pdf} & \includegraphics[scale=0.2]{figures/dots/4.pdf} & \includegraphics[scale=0.2]{figures/dots/1.pdf} & \includegraphics[scale=0.2]{figures/dots/1.pdf} & \includegraphics[scale=0.2]{figures/dots/1.pdf} & \includegraphics[scale=0.2]{figures/dots/1.pdf}  & \includegraphics[scale=0.2]{figures/dots/1.pdf} &—&—& \includegraphics[scale=0.2]{figures/dots/1.pdf} & \includegraphics[scale=0.2]{figures/dots/1.pdf} \\ 
        
         & TSIQA~\cite{xiaoModelingTemporalSensitiveInformation2022a} & \includegraphics[scale=0.2]{figures/dots/1.pdf} &—&— & \includegraphics[scale=0.2]{figures/dots/1.pdf} & \includegraphics[scale=0.2]{figures/dots/1.pdf} & \includegraphics[scale=0.2]{figures/dots/1.pdf} & \includegraphics[scale=0.2]{figures/dots/4.pdf} & \includegraphics[scale=0.2]{figures/dots/4.pdf} & \includegraphics[scale=0.2]{figures/dots/4.pdf} &\includegraphics[scale=0.2]{figures/dots/4.pdf} & \includegraphics[scale=0.2]{figures/dots/1.pdf} & \includegraphics[scale=0.2]{figures/dots/1.pdf}  & \includegraphics[scale=0.2]{figures/dots/1.pdf} &—&—  & \includegraphics[scale=0.2]{figures/dots/1.pdf} & \includegraphics[scale=0.2]{figures/dots/4.pdf} \\ 
        
         & TMA~\cite{liuTimeAwareMultiwayAdaptive2023}  & \includegraphics[scale=0.2]{figures/dots/1.pdf} &—&—& \includegraphics[scale=0.2]{figures/dots/1.pdf} & \includegraphics[scale=0.2]{figures/dots/1.pdf} & \includegraphics[scale=0.2]{figures/dots/4.pdf} & \includegraphics[scale=0.2]{figures/dots/4.pdf} & \includegraphics[scale=0.2]{figures/dots/4.pdf} & \includegraphics[scale=0.2]{figures/dots/1.pdf} & \includegraphics[scale=0.2]{figures/dots/4.pdf} & \includegraphics[scale=0.2]{figures/dots/1.pdf} & \includegraphics[scale=0.2]{figures/dots/1.pdf}  & \includegraphics[scale=0.2]{figures/dots/1.pdf} &—&— & \includegraphics[scale=0.2]{figures/dots/1.pdf} & \includegraphics[scale=0.2]{figures/dots/1.pdf} \\ 
        
         & MultiQA~\cite{chenMultigranularityTemporalQuestion2023} & \includegraphics[scale=0.2]{figures/dots/4.pdf} & \includegraphics[scale=0.2]{figures/dots/4.pdf} & \includegraphics[scale=0.2]{figures/dots/4.pdf} &  \includegraphics[scale=0.2]{figures/dots/1.pdf} & \includegraphics[scale=0.2]{figures/dots/1.pdf} & \includegraphics[scale=0.2]{figures/dots/1.pdf} & \includegraphics[scale=0.2]{figures/dots/1.pdf} & \includegraphics[scale=0.2]{figures/dots/1.pdf} & \includegraphics[scale=0.2]{figures/dots/1.pdf} & \includegraphics[scale=0.2]{figures/dots/1.pdf} & \includegraphics[scale=0.2]{figures/dots/1.pdf} & \includegraphics[scale=0.2]{figures/dots/1.pdf}  & \includegraphics[scale=0.2]{figures/dots/4.pdf} & \includegraphics[scale=0.2]{figures/dots/4.pdf} & \includegraphics[scale=0.2]{figures/dots/4.pdf}  & \includegraphics[scale=0.2]{figures/dots/1.pdf} & \includegraphics[scale=0.2]{figures/dots/1.pdf} \\ 
        
         & LGQA~\cite{liuLocalGlobalTemporal2023} & \includegraphics[scale=0.2]{figures/dots/1.pdf} & \includegraphics[scale=0.2]{figures/dots/1.pdf} & \includegraphics[scale=0.2]{figures/dots/1.pdf} & \includegraphics[scale=0.2]{figures/dots/1.pdf} & \includegraphics[scale=0.2]{figures/dots/1.pdf} & \includegraphics[scale=0.2]{figures/dots/1.pdf} & \includegraphics[scale=0.2]{figures/dots/4.pdf} & \includegraphics[scale=0.2]{figures/dots/1.pdf} & \includegraphics[scale=0.2]{figures/dots/1.pdf} & \includegraphics[scale=0.2]{figures/dots/4.pdf} & \includegraphics[scale=0.2]{figures/dots/1.pdf} & \includegraphics[scale=0.2]{figures/dots/1.pdf}  & \includegraphics[scale=0.2]{figures/dots/1.pdf} & \includegraphics[scale=0.2]{figures/dots/1.pdf} & \includegraphics[scale=0.2]{figures/dots/1.pdf}  & \includegraphics[scale=0.2]{figures/dots/1.pdf} & \includegraphics[scale=0.2]{figures/dots/4.pdf} \\ 
        
         & JMFRN~\cite{huangJointMultiFactsReasoning2024} & \includegraphics[scale=0.2]{figures/dots/1.pdf} &—&—&  \includegraphics[scale=0.2]{figures/dots/1.pdf} & \includegraphics[scale=0.2]{figures/dots/4.pdf} & \includegraphics[scale=0.2]{figures/dots/1.pdf} & \includegraphics[scale=0.2]{figures/dots/1.pdf} & \includegraphics[scale=0.2]{figures/dots/4.pdf} & \includegraphics[scale=0.2]{figures/dots/1.pdf} & \includegraphics[scale=0.2]{figures/dots/1.pdf} & \includegraphics[scale=0.2]{figures/dots/1.pdf} & \includegraphics[scale=0.2]{figures/dots/1.pdf}  & \includegraphics[scale=0.2]{figures/dots/1.pdf} &—& — & \includegraphics[scale=0.2]{figures/dots/1.pdf} & \includegraphics[scale=0.2]{figures/dots/4.pdf} \\ 
        
         & SERQA~\cite{duSemanticenhancedReasoningQuestion2024} & \includegraphics[scale=0.2]{figures/dots/1.pdf} & \includegraphics[scale=0.2]{figures/dots/1.pdf} & \includegraphics[scale=0.2]{figures/dots/1.pdf} &  \includegraphics[scale=0.2]{figures/dots/1.pdf} & \includegraphics[scale=0.2]{figures/dots/1.pdf} & \includegraphics[scale=0.2]{figures/dots/1.pdf} & \includegraphics[scale=0.2]{figures/dots/4.pdf} & \includegraphics[scale=0.2]{figures/dots/4.pdf} & \includegraphics[scale=0.2]{figures/dots/1.pdf} & \includegraphics[scale=0.2]{figures/dots/1.pdf} & \includegraphics[scale=0.2]{figures/dots/1.pdf} & \includegraphics[scale=0.2]{figures/dots/1.pdf}  & \includegraphics[scale=0.2]{figures/dots/1.pdf} & \includegraphics[scale=0.2]{figures/dots/1.pdf} & \includegraphics[scale=0.2]{figures/dots/1.pdf}  & \includegraphics[scale=0.2]{figures/dots/1.pdf} & \includegraphics[scale=0.2]{figures/dots/4.pdf} \\ 
        
         & QC-MHM~\cite{xueQuestionCalibrationMultiHop2024}  & \includegraphics[scale=0.2]{figures/dots/1.pdf} &—&—& \includegraphics[scale=0.2]{figures/dots/4.pdf} & \includegraphics[scale=0.2]{figures/dots/4.pdf} & \includegraphics[scale=0.2]{figures/dots/4.pdf} & \includegraphics[scale=0.2]{figures/dots/4.pdf} & \includegraphics[scale=0.2]{figures/dots/4.pdf} & \includegraphics[scale=0.2]{figures/dots/1.pdf} & \includegraphics[scale=0.2]{figures/dots/1.pdf} & \includegraphics[scale=0.2]{figures/dots/1.pdf} & \includegraphics[scale=0.2]{figures/dots/1.pdf}  & \includegraphics[scale=0.2]{figures/dots/4.pdf} &—& — & \includegraphics[scale=0.2]{figures/dots/1.pdf} & \includegraphics[scale=0.2]{figures/dots/1.pdf} \\


         & $M^3$TQA~\cite{zhaM3TQAMultiViewMultiHop2024a}  & \includegraphics[scale=0.2]{figures/dots/1.pdf} & \includegraphics[scale=0.2]{figures/dots/1.pdf} & \includegraphics[scale=0.2]{figures/dots/1.pdf} & \includegraphics[scale=0.2]{figures/dots/4.pdf} & \includegraphics[scale=0.2]{figures/dots/4.pdf} & \includegraphics[scale=0.2]{figures/dots/1.pdf} & \includegraphics[scale=0.2]{figures/dots/1.pdf} & \includegraphics[scale=0.2]{figures/dots/1.pdf} & \includegraphics[scale=0.2]{figures/dots/1.pdf} & \includegraphics[scale=0.2]{figures/dots/1.pdf} & \includegraphics[scale=0.2]{figures/dots/1.pdf} &  \includegraphics[scale=0.2]{figures/dots/1.pdf}  & \includegraphics[scale=0.2]{figures/dots/4.pdf} & \includegraphics[scale=0.2]{figures/dots/4.pdf} & \includegraphics[scale=0.2]{figures/dots/4.pdf}  & \includegraphics[scale=0.2]{figures/dots/1.pdf} & \includegraphics[scale=0.2]{figures/dots/4.pdf} \\  

         & MusTKGQA~\cite{zhangMusTQTemporalKnowledge2024}  & \includegraphics[scale=0.2]{figures/dots/1.pdf} & \includegraphics[scale=0.2]{figures/dots/1.pdf} & \includegraphics[scale=0.2]{figures/dots/1.pdf} & \includegraphics[scale=0.2]{figures/dots/1.pdf} & \includegraphics[scale=0.2]{figures/dots/1.pdf} & \includegraphics[scale=0.2]{figures/dots/1.pdf} & \includegraphics[scale=0.2]{figures/dots/1.pdf} & \includegraphics[scale=0.2]{figures/dots/1.pdf} & \includegraphics[scale=0.2]{figures/dots/1.pdf} & \includegraphics[scale=0.2]{figures/dots/1.pdf} & \includegraphics[scale=0.2]{figures/dots/1.pdf} & \includegraphics[scale=0.2]{figures/dots/1.pdf}  & \includegraphics[scale=0.2]{figures/dots/1.pdf} & \includegraphics[scale=0.2]{figures/dots/1.pdf} & \includegraphics[scale=0.2]{figures/dots/1.pdf}  & \includegraphics[scale=0.2]{figures/dots/1.pdf} & \includegraphics[scale=0.2]{figures/dots/1.pdf} \\ \midrule[1 pt]
        
    \multirow{6}{*}{\shortstack{LLM-based}} 
    & Prog-TQA~\cite{chenSelfImprovementProgrammingTemporal2024} & \includegraphics[scale=0.2]{figures/dots/4.pdf} & \includegraphics[scale=0.2]{figures/dots/4.pdf} & \includegraphics[scale=0.2]{figures/dots/4.pdf} & \includegraphics[scale=0.2]{figures/dots/1.pdf} & \includegraphics[scale=0.2]{figures/dots/1.pdf} & \includegraphics[scale=0.2]{figures/dots/4.pdf} & \includegraphics[scale=0.2]{figures/dots/4.pdf} & \includegraphics[scale=0.2]{figures/dots/4.pdf} & \includegraphics[scale=0.2]{figures/dots/4.pdf} & \includegraphics[scale=0.2]{figures/dots/4.pdf} & \includegraphics[scale=0.2]{figures/dots/4.pdf} & \includegraphics[scale=0.2]{figures/dots/1.pdf}  & \includegraphics[scale=0.2]{figures/dots/4.pdf} & \includegraphics[scale=0.2]{figures/dots/4.pdf} & \includegraphics[scale=0.2]{figures/dots/4.pdf}  & \includegraphics[scale=0.2]{figures/dots/1.pdf} & \includegraphics[scale=0.2]{figures/dots/4.pdf} \\
    
     & ARI~\cite{chenTemporalKnowledgeQuestion2023} & \includegraphics[scale=0.2]{figures/dots/1.pdf} & \includegraphics[scale=0.2]{figures/dots/1.pdf} & \includegraphics[scale=0.2]{figures/dots/1.pdf} & \includegraphics[scale=0.2]{figures/dots/1.pdf} & \includegraphics[scale=0.2]{figures/dots/1.pdf} & \includegraphics[scale=0.2]{figures/dots/1.pdf} & \includegraphics[scale=0.2]{figures/dots/1.pdf} & \includegraphics[scale=0.2]{figures/dots/1.pdf} & \includegraphics[scale=0.2]{figures/dots/1.pdf} & \includegraphics[scale=0.2]{figures/dots/4.pdf} & \includegraphics[scale=0.2]{figures/dots/1.pdf} &—& \includegraphics[scale=0.2]{figures/dots/1.pdf} & \includegraphics[scale=0.2]{figures/dots/1.pdf} & \includegraphics[scale=0.2]{figures/dots/1.pdf} & \includegraphics[scale=0.2]{figures/dots/1.pdf} & \includegraphics[scale=0.2]{figures/dots/4.pdf} \\      

    & QAap~\cite{zhuQuestionAnsweringProgramming2023} & \includegraphics[scale=0.2]{figures/dots/1.pdf} & \includegraphics[scale=0.2]{figures/dots/1.pdf} & \includegraphics[scale=0.2]{figures/dots/1.pdf} & \includegraphics[scale=0.2]{figures/dots/4.pdf} & \includegraphics[scale=0.2]{figures/dots/1.pdf} & \includegraphics[scale=0.2]{figures/dots/1.pdf} & \includegraphics[scale=0.2]{figures/dots/1.pdf} & \includegraphics[scale=0.2]{figures/dots/1.pdf} & — & — & — &  \includegraphics[scale=0.2]{figures/dots/1.pdf} & \includegraphics[scale=0.2]{figures/dots/1.pdf} & \includegraphics[scale=0.2]{figures/dots/1.pdf} & \includegraphics[scale=0.2]{figures/dots/1.pdf} & \includegraphics[scale=0.2]{figures/dots/1.pdf} & \includegraphics[scale=0.2]{figures/dots/1.pdf} \\  
    
     & GenTKGQA~\cite{gaoTwostageGenerativeQuestion2024} & \includegraphics[scale=0.2]{figures/dots/1.pdf} &—&—& \includegraphics[scale=0.2]{figures/dots/1.pdf} & \includegraphics[scale=0.2]{figures/dots/1.pdf} & \includegraphics[scale=0.2]{figures/dots/4.pdf} & \includegraphics[scale=0.2]{figures/dots/4.pdf} & \includegraphics[scale=0.2]{figures/dots/4.pdf} & \includegraphics[scale=0.2]{figures/dots/1.pdf} & \includegraphics[scale=0.2]{figures/dots/1.pdf} & \includegraphics[scale=0.2]{figures/dots/1.pdf} & \includegraphics[scale=0.2]{figures/dots/1.pdf} & \includegraphics[scale=0.2]{figures/dots/4.pdf} &—& — & \includegraphics[scale=0.2]{figures/dots/4.pdf} & \includegraphics[scale=0.2]{figures/dots/4.pdf} \\  
     
     & FAITH~\cite{jiaFaithfulTemporalQuestion2024}  & \includegraphics[scale=0.2]{figures/dots/1.pdf} & \includegraphics[scale=0.2]{figures/dots/1.pdf} & \includegraphics[scale=0.2]{figures/dots/1.pdf} & \includegraphics[scale=0.2]{figures/dots/1.pdf} & \includegraphics[scale=0.2]{figures/dots/4.pdf} & \includegraphics[scale=0.2]{figures/dots/1.pdf} & \includegraphics[scale=0.2]{figures/dots/1.pdf} & \includegraphics[scale=0.2]{figures/dots/1.pdf} & \includegraphics[scale=0.2]{figures/dots/1.pdf} &\includegraphics[scale=0.2]{figures/dots/1.pdf} & \includegraphics[scale=0.2]{figures/dots/1.pdf} & \includegraphics[scale=0.2]{figures/dots/1.pdf}  & \includegraphics[scale=0.2]{figures/dots/1.pdf} & \includegraphics[scale=0.2]{figures/dots/1.pdf} & \includegraphics[scale=0.2]{figures/dots/1.pdf}  & \includegraphics[scale=0.2]{figures/dots/1.pdf} & \includegraphics[scale=0.2]{figures/dots/1.pdf} \\ 
    
    & Time$R^4$~\cite{qianTimeR4TimeawareRetrievalAugmented2024} & \includegraphics[scale=0.2]{figures/dots/1.pdf} & \includegraphics[scale=0.2]{figures/dots/1.pdf} & \includegraphics[scale=0.2]{figures/dots/1.pdf} & \includegraphics[scale=0.2]{figures/dots/1.pdf} & \includegraphics[scale=0.2]{figures/dots/1.pdf} & \includegraphics[scale=0.2]{figures/dots/1.pdf} & \includegraphics[scale=0.2]{figures/dots/1.pdf} & \includegraphics[scale=0.2]{figures/dots/4.pdf} & \includegraphics[scale=0.2]{figures/dots/1.pdf} & \includegraphics[scale=0.2]{figures/dots/1.pdf} & \includegraphics[scale=0.2]{figures/dots/1.pdf} &  \includegraphics[scale=0.2]{figures/dots/4.pdf} & \includegraphics[scale=0.2]{figures/dots/4.pdf} & \includegraphics[scale=0.2]{figures/dots/4.pdf} & \includegraphics[scale=0.2]{figures/dots/4.pdf} & \includegraphics[scale=0.2]{figures/dots/1.pdf} & \includegraphics[scale=0.2]{figures/dots/1.pdf} \\
    \bottomrule[1 pt]
    \end{NiceTabular}
\end{table*}

\subsection{Comparison Across TKGQA Methods}
\label{taskcoverage}

Building on the question taxonomy and methodologies overview, we match each type of temporal question with the appropriate method to address it effectively, providing a detailed table as Table~\ref{align}. According to the model design and experimental results, we categorize the model in correspondence with the question types into the following categories: i) Model design specifically addresses a certain type of question (darker blue); ii) Questions that the model can solve (light blue); iii) Questions that the model cannot solve or has not considered (dash).

Based on the table, we have the following observations:
Semantic parsing-based approaches demonstrate comprehensive performance in all temporal granularities (Year/Month/Day) and temporal expressions (Explicit/Implicit), achieving near-complete ``solvable'' status. This is because, in the logical form, it is easy to design different operators for discrete-time granularity. It is straightforward to use temporal operators to transform implicit expressions into explicit ones, though these explicit forms can be challenging to implement efficiently in embedding calculations.
However, they focus less on complex questions, which may stem from the increased labour costs and technical complexity associated with designing specialised representation forms for more sophisticated temporal reasoning.

TKG Embedding-based methods focus more strongly on temporal signals (e.g., Overlap, Before/After) and complex temporal questions, but demonstrate weaker temporal granularity coverage (primarily supporting Year-level resolution). The advantage of the TKG Embedding-based approach is that it can design evaluation mechanisms tailored to different question types or employ techniques like contrastive learning to highlight specific temporal signals. However, because TKG timestamps are fixed to a single granularity, such as years, the TKG representation model learns only one-time granularity, making it challenging to address questions requiring finer or coarser granularity. These methods also prioritise model design for answer prediction due to the need for specialised architectures for different answer types.

LLM-based methods maintain suboptimal coverage scope. Certain methods demonstrate significant expertise in processing complex temporal signals.
Complex temporal problems receive significant attention from most approaches, reflecting their critical role in temporal question answering.
There is an increased focus on multi-granularity temporal support (e.g., MultiQA, LGQA) and complex temporal reasoning (e.g., JMFRN, QC-MHM) in recent advancements.

\begin{figure*}[htbp]
  \centering
  \subfloat[Question Answering on Cron questions (upper), ComplexCronQuestions (middle) and MultiTQ (lower).]{\includegraphics[width=0.515\textwidth]{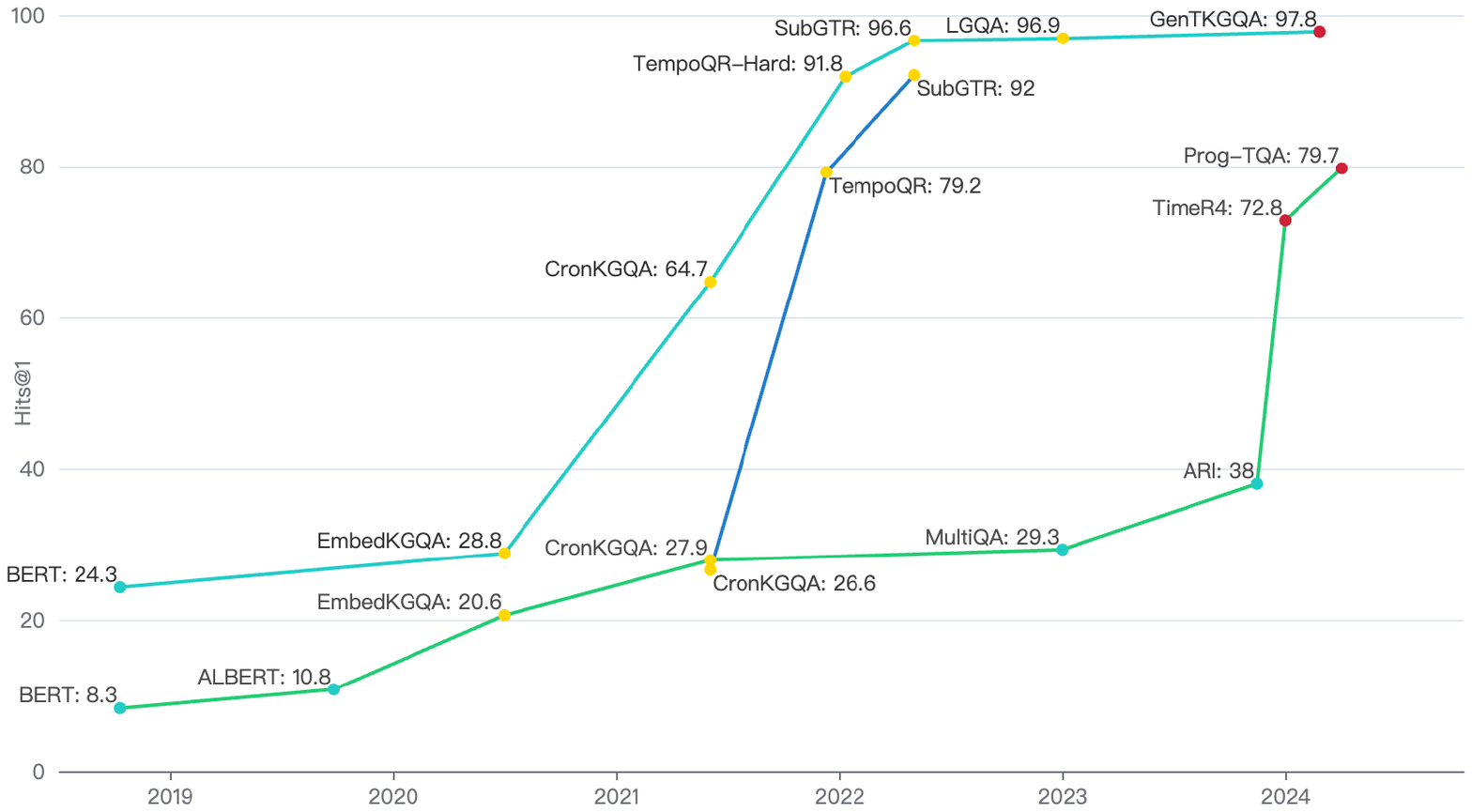}}
  \subfloat[Question Answering on TempQuestions (upper) and TempQA-WD (lower).]{\includegraphics[width=0.485\textwidth]{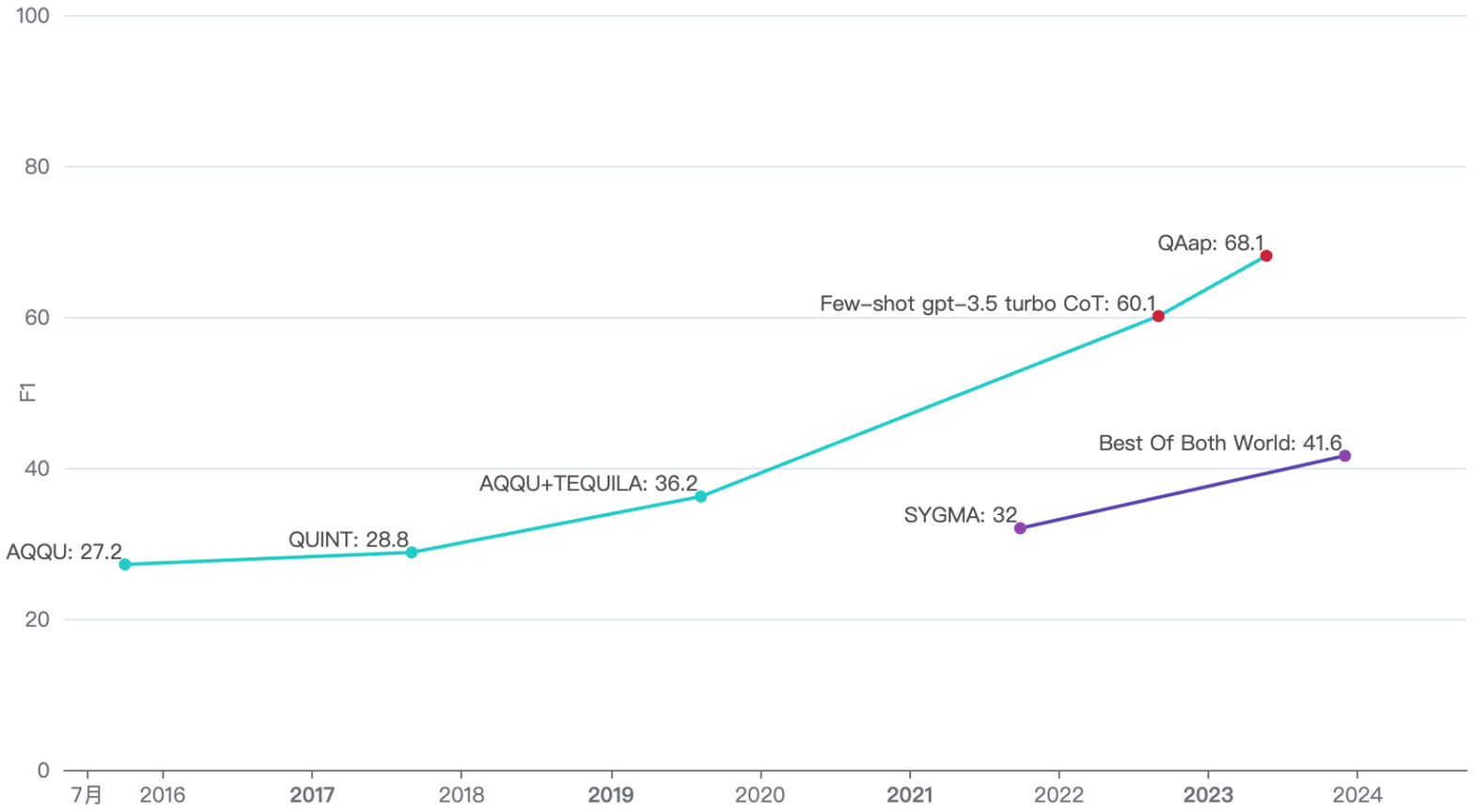}}
    \hfill
  \subfloat[Question Answering on TimeQuestions (upper) and TIQ (lower).]{\includegraphics[width=0.5\textwidth]{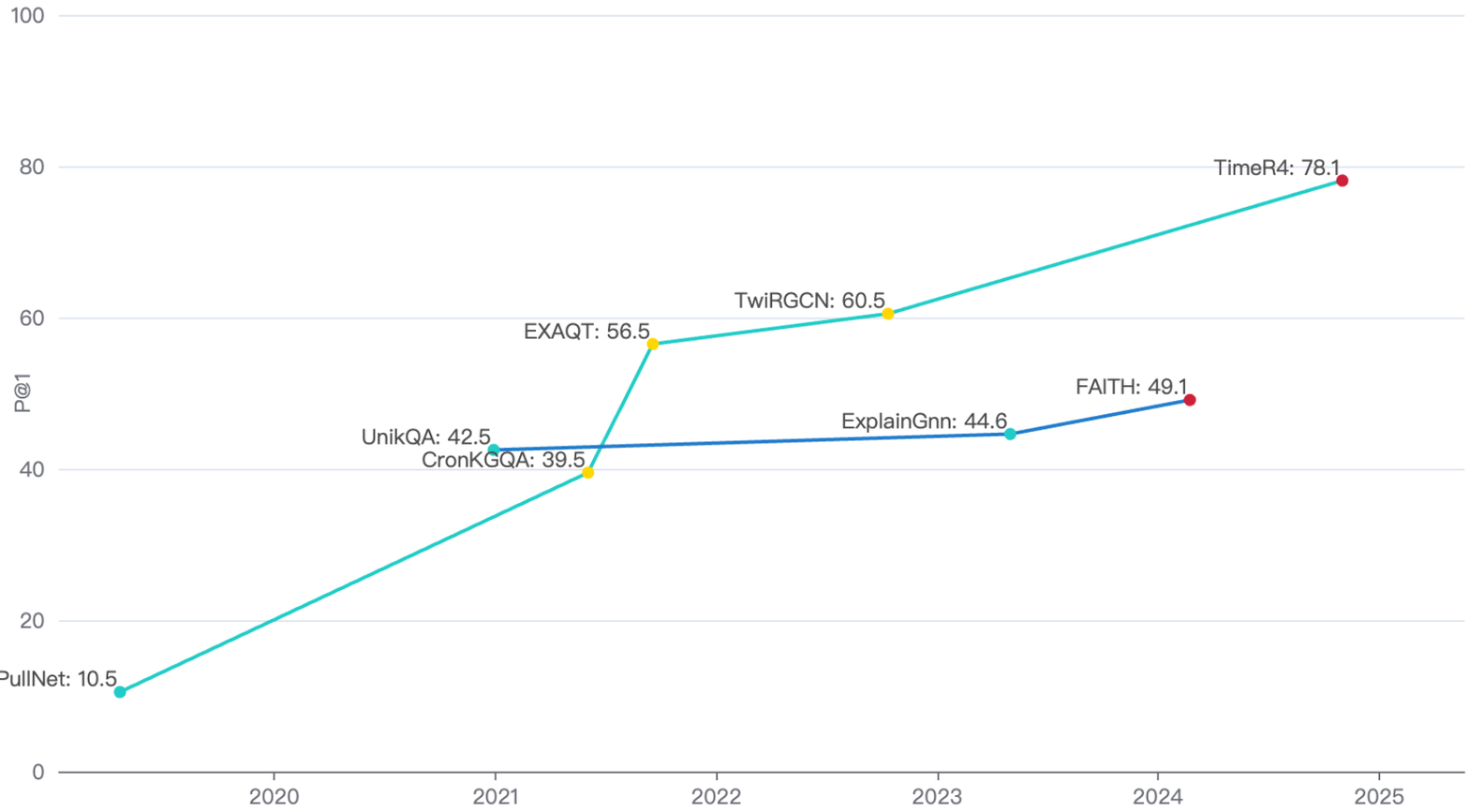}}

\caption{Leaderboard on TKGQA datasets. The purple dots represent SP-based methods. The yellow dots represent TKG Embedding-based methods. The red dots represent LLM-based methods. }
  \label{fig:leaderboard}
\end{figure*}

\subsection{Evaluation Metrics and Leaderboard}
This section explains the evaluation metrics for the TKGQA task and provides a leaderboard to compare the performance of various methods across different datasets, enabling a comprehensive quantitative comparison of existing methods.

\subsubsection{Evaluation Metrics}
\label{sec-metrics}

\noindent{\textbf{Hits@k.}} This is the most widely used metric for the TKGQA task, where $k = 1,3,5,10$. The metric is set to one if a correct answer appears within the top $k$ positions and zero otherwise.

\noindent{\textbf{Precision, Recall and F1 score.}} These metrics are widely used for the KGQA task. Precision indicates the ratio of the correct predictions to all the predicted answers. 
Recall is the ratio of the correct predictions to the ground truth.
The F1 score computes the average of precision and recall.
\[
F_1 = 2 \ast \frac{\text{Precision} \ast \text{Recall}}{\text{Precision} + \text{Recall}}.
\]

\noindent{\textbf{P@1.}} Precision at the top rank is one if the highest ranked answer is correct and zero otherwise.

\noindent{\textbf{MRR.}} This is the reciprocal of the first rank, where we have a correct answer. If the correct answer is not featured in the ranked list, MRR is zero.

\noindent{\textbf{Average number of reasoning steps.}} 
ARI uses this metric to measure the reasoning steps for each question. The average number of reasoning steps across all tested questions represents this metric.
\subsubsection{Leaderboard}
\label{leaderboard}
Figure \ref{fig:leaderboard} presents leaderboards across the mentioned datasets. Detailed descriptions of datasets are shown in Appendix \ref{sec-datasets}. 

Based on the leaderboards, we have the following observations:
\begin{inparaenum}[(i)]
\item LLM-based methods dominate recent benchmarks. In CronQuestions, TimeQuestions, MultiTQ, TempQuestions, LLM-based methods achieve the SOTA result, demonstrating their superiority;
\item  TKGE-based approaches exhibit broader applicability. TKGE-based models consistently rank high across multiple datasets (2018–2024). Meanwhile, SP-based methods are less prevalent, possibly due to limitations in handling dynamic temporal patterns and high demand for the dataset;
\item Significant gaps persist in underperforming datasets. For example, TimeQuestion's top model (TimeR$^4$) achieves only 78.1\% P@1, and TIQ's leading model (Unit-Qa) attains 60\% P@1. These results highlight unresolved challenges in tasks requiring fine-grained temporal alignment or open-domain answer generation.
\end{inparaenum}
These observations underscore the rapid evolution of TKGQA methods and architectures driving progress, while specific tasks demand further innovation.

\section{Future Directions\label{future}}
This section will discuss emerging frontiers for TKGQA,  aiming to stimulate further research in this field.

\subsection{More Question Types}
While existing datasets already cover some of the temporal questions, there are still more questions to be explored in the real world. 
\begin{inparaenum}[(i)]
\item  More combination of existing question types: ``Who was the \textbf{first} person to win a medal \textbf{during} the 2024 Olympic Games?'' 
\item More time granularity: Some questions demand more fine-grained granularities, such as ``When was the Long March 1 launched?'' 
\item Questions must consider the posed time: ``Where are the Seneca Indians now?''~\cite{jiaComplexTemporalQuestion2021,liskaStreamingQABenchmarkAdaptation2022a} 
\item  Predicting the future questions: ``Will the Palestinian-Israeli conflict end next year?''~\cite{jinForecastQAQuestionAnswering2021, dingForecastingQuestionAnswering2022, dingForecastTKGQuestionsBenchmarkTemporal2022} \item Common sense temporal questions: ``How often are the Olympics held?''
\end{inparaenum}

\subsection{Robust Model}
The current TKGQA methods are mostly based on the gold annotated entities and temporal signals given in the dataset~\cite{saxenaQuestionAnsweringTemporal2021,jiaComplexTemporalQuestion2021, neelamBenchmarkGeneralizableInterpretable2022}. However, these models lack robustness against real-world data, which stems from two aspects: on the one hand, the annotated entities and temporal signals in real-world data may contain errors, and the models need to have fault tolerance; on the other hand, the models need to have the ability to generalize to unseen entities and temporal signals in the training set. These limitations can be addressed in future work.

\subsection{Multi-modal TKGQA} Current TKGQA systems mainly handle plain text input. However, we experience the world with multiple modalities (e.g., video). Therefore, building a multi-modal TKGQA system that can handle multiple modalities is an important direction to investigate~\cite{yuProphetPromptingLarge2023}. 
A non-trivial challenge is effectively making a multimodal feature alignment, and complementary to better understand the temporal part.
\subsection{LLM for TKGQA} Recently, LLMs have gained significant attention for their remarkable performance across a wide range of Natural Language Processing (NLP) tasks~\cite{touvronLlamaOpenFoundation,openaiGPT4TechnicalReport2024,geminiteamGeminiFamilyHighly2024}. Existing research has also explored applying LLMs in KGQA scenarios, employing both few-shot and zero-shot learning paradigms~\cite{nieCodeStyleContextLearning2024,sunThinkGraphDeepResponsible2024, jiangStructGPTGeneralFramework2023a,baekKnowledgeAugmentedLanguageModel2023,liFewshotIncontextLearning2023a,liUnlockingTemporalQuestion2023a}
    Several emerging opportunities could further enhance the capabilities of LLMs in TKGQA systems:
    \begin{enumerate}
        \item \textbf{Multi-Agent Collaboration Interactive Reasoning for TKGQA.} 
    Recent LLM works have shifted the focus from traditional NLP tasks to exploring language agents in simulation environments that mimic real-world scenarios~\cite{zhangTimeArenaShapingEfficient2024}. Qian~\cite{qianScalingLargeLanguageModelbasedMultiAgent2024} investigates interactive reasoning and collective intelligence in autonomously solving complex
    problems. This may be further explored for temporal reasoning in temporal questions.    
    \item \textbf{Diverse Data Generation.} Numerous studies have demonstrated the effectiveness of large models in data generation~\cite{chungIncreasingDiversityMaintaining2023}, which can be used to enhance the diversity of the TKGQA dataset.
    \item \textbf{Supplementing Knowledge.} The language model itself can serve as a TKG as demonstrated by Dhingra~\cite{dhingraTimeAwareLanguageModels2022}. Additionally, LLMs possess temporal commonsense~\cite{chuTimeBenchComprehensiveEvaluation2023}, which is often absent in traditional temporal knowledge graphs. This temporal knowledge can complement existing TKGs for TKGQA.
    \end{enumerate}

\section{Conclusion\label{conclusion}}
In this paper, we provided an in-depth analysis of the emerging field of TKGQA with a new taxonomy of temporal questions and a systematic categorization of existing methods. We demonstrated the focus and neglect of existing methods for temporal questions, indicating future research directions. We have discussed some new trends in this research field, hoping to attract more breakthroughs in future research.



\balance

\bibliographystyle{IEEEtran}
\bibliography{IEEEabrv, sections/main, sections/LLM,sections/TKG}

\appendices

\section{Datasets}
\label{sec-datasets}
This section introduces the datasets, including their background TKG, size, etc. We provide a question category coverage comparison across TKGQA datasets in Table~\ref{questincategoryfordataset}.

\subsection{TempQuestions} TempQuestions \cite{jiaTempQuestionsBenchmarkTemporal2018} is a benchmark dataset derived from Freebase~\cite{bollackerFreebaseCollaborativelyCreated2008}, where temporal knowledge is stored using compound value types (CVTs). Examples of these CVTs include \texttt{footballPlayer.team.joinedOnDate}, \texttt{footballPlayer.team.leftOnDate},
\texttt{marriage.date},
\texttt{amusement\_parks.ride.opened}	,\texttt{amusement\_parks.ride.closed}.
It includes 1,271 questions with temporal signals, question types, and data sources for testing and evaluation.

\subsection{\textbf{TempQA-WD}} TempQA-WD~\cite{neelamSYGMASystemGeneralizable2021} 
is an adaptation of TempQuestions for Wikidata~\cite{vrandecicWikidataFreeCollaborative2014}, fulfilling the need for a multi-KG-based dataset to evaluate their model.
The dataset comprises:
\begin{itemize}
    \item 839 questions with corresponding Wikidata SPARQL queries, answers, categories, and TempQuestions' information.
    \item 175 questions with AMR, lambda expressions, entities, relations, and KB-specific lambda expressions, in addition to the above information.
\end{itemize}

\subsection{\textbf{TimeQuestions}}TimeQuestions~\cite{jiaComplexTemporalQuestion2021}
is based on Wikidata, it includes temporal facts of triples, such as 
\texttt{<Malia Obama, date of birth, 04-07-1998>} or maintain more temporal knowledge with qualifiers like
\texttt{<Barack Obama, position held, President of the US; start date, 20-01-2009; end date, 20-01-2017>}. 
Jia searched through eight KG-QA datasets for time-related questions and mapped them to Wikidata. Questions in each benchmark are tagged for temporal expressions using SUTime~\cite{changSUTIMELibraryRecognizing} and HeidelTime~\cite{strotgenHeidelTimeHighQuality2010} and for signal words using a dictionary compiled by Setzer~\cite{setzerTemporalInformationNewswire2001a} and manually tagged with its temporal question category.
In total, the TimeQuestions comprises 16,859 questions.

\subsection{\textbf{CronQuestions}}CronQuestions~\cite{chenTemporalKnowledgeGraph2022} utilises a subset of Wikidata that includes facts annotated with temporal information~\cite{lacroixTensorDecompositionsTemporal2020}, such as \texttt{<Barack Obama, held position, President of USA, 2008, 2016>}. Entities extracted from Wikidata with both ``start time'' and ``end time'' annotations are transformed into event format (e.g., \texttt{<WWII, significant event, occurred, 1939, 1945>}). 
The dataset comprises a Temporal KG with 125k entities and 328k facts (including 5k event facts), and 410k natural language questions requiring temporal reasoning.

\subsection{\textbf{Complex-CronQuestions}}
Chen~\cite{chenTemporalKnowledgeGraph2022} observe that existing benchmarks contain many pseudo-temporal questions. For instance, for the question \textit{``What’s the first award Carlo Taverna got?''} There is only one fact related to\textit{ Carlo Taverna} in the TKG, which makes the temporal word ``first'' meaningless as a constraint. 
They remove all simple and pseudo-temporal questions and filter out questions with fewer than 5 relevant facts in CronQuestions.

\subsection{\textbf{MultiTQ}}MultiTQ \cite{chenMultigranularityTemporalQuestion2023} is a dataset derived from ICEWS05-15~\cite{garcia-duranLearningSequenceEncoders2018}, where all facts are standardized as quadruple $(s,r,o,t)$.
ICEWS05-15 is notable for its rich semantic information, with a higher average number of relation types per entity than other TKGs.
The MultiTQ dataset features several advantages, including its large scale, ample relations, and multiple temporal granularities.
ICEWS provides time information at a day granularity, while the authors generate higher granularities, such as year and month, for the questions. MultiTQ contains 500,000 questions, making it a significant resource for temporal question-answering research.

\subsection{\textbf{MusTQ}}MusTQ \cite{zhangMusTQTemporalKnowledge2024} is a dataset that contains diverse multi-step temporal reasoning. MusTQ contains ~666K temporal questions and a WikiData subset (with 125k entities and 326k facts) as its knowledge base. It has six question types and four answer types (entity, time, boolean, numeric), and the maximum number of reasoning steps required for the questions is up to four.

\subsection{\textbf{ForecastTKGQuestions}}ForecastTKGQuestions \cite{dingForecastTKGQuestionsBenchmarkTemporal2022} aims to predict facts in the future timestamps beyond the ICEWS subset. It includes three types of forecasting questions, i.e., entity prediction, yes-unknown, and fact reasoning questions. The answer types for each category are entity, yes-unknown label, and choice label, respectively. The total number of questions in ForecastTKGQuestions is up to 727k. Entities and timestamps are annotated in their questions. 

\subsection{\textbf{TIQ}}TIQ~\cite{jiaFaithfulTemporalQuestion2024} is a benchmark with a primary focus on challenging and diverse implicit questions, and it has multiple sources: Wikipedia text and infoboxes, and the Wikidata KB. It has 10,000 questions with metadata. The metadata includes the information snippets grounding the question, the sources these were obtained, the explicit temporal value expressed by the implicit constraint, the topic entity, the question entities detected in the snippets, and the temporal signal.


\begin{table*}[htbp]
    \caption{Question category coverage comparison across TKGQA datasets.\label{questincategoryfordataset}}
    \centering
    \scriptsize
    \renewcommand{\arraystretch}{1.8}
    \resizebox{\linewidth}{!}{
    \begin{NiceTabular}{|c|c|c|c|c|c|c|c|c|c|}
    \toprule
        \Block{1-3}{\textbf{Category}}& & & \multicolumn{1}{m{1.1cm}}{\textbf{\shortstack{Temp\\Questions}}} & \multicolumn{1}{m{1.1cm}}{\textbf{\shortstack{TempQA-\\WD}}} & \multicolumn{1}{m{1.1cm}}{\textbf{\shortstack{Time\\Questions}}} & \multicolumn{1}{m{1.1cm}}{\textbf{\shortstack{Cron\\Questions}}} & \multicolumn{1}{m{1.6cm}}{\textbf{\shortstack{Complex\\CronQuestions}}} & \textbf{MultiQA} & \textbf{MusTQ}\\ \hline
        \multirow{13}{*}{Question Content} & \multirow{3}{*}{\shortstack{Time\\Granularity}} & Year & \checkmark & \checkmark & \checkmark & \checkmark &\checkmark& \checkmark &\\ 
        ~ & ~ & Month & \checkmark & \checkmark & ~ & ~ & & \checkmark & \checkmark \\
        ~ & ~ & Day & \checkmark & \checkmark & \checkmark & ~ && \checkmark & \\  \cline{2-10}
        ~ & \multirow{2}{*}{\shortstack{Time\\Expression}} & Explicit & \checkmark & \checkmark & \checkmark & \checkmark &\checkmark& \checkmark & \\ 
        ~ & ~ & Implicit & \checkmark & \checkmark & \checkmark & \checkmark & \checkmark&\checkmark & \checkmark \\ \cline{2-10}
        ~ & \multirow{6}{*}{\shortstack{Temporal\\Signal}} & Overlap & \checkmark & \checkmark & \checkmark & \checkmark &\checkmark &\checkmark & \checkmark \\ 
        ~ & ~ & Equal & \checkmark & \checkmark & \checkmark & \checkmark &\checkmark &\checkmark & \checkmark \\ 
        ~ & ~ & Start/End & \checkmark & \checkmark & \checkmark & \checkmark &\checkmark & ~ & \\ 
        ~ & ~ & During/Include & \checkmark & \checkmark & \checkmark & \checkmark &\checkmark & \checkmark & \checkmark\\ 
        ~ & ~ & Before/After & \checkmark & \checkmark & \checkmark & \checkmark &\checkmark &\checkmark & \checkmark\\ 
        ~ & ~ & Ordinal & \checkmark & \checkmark & \checkmark & \checkmark & \checkmark&\checkmark & \checkmark\\ \cline{2-10}
        ~ & {\multirow{2}{*}{\shortstack{Temporal\\Constraint\\Composition}}} & w/ Composition & & ~ & ~ & ~ & &\checkmark & \checkmark\\
        ~ & ~ & w/o Composition & \checkmark & \checkmark & \checkmark & \checkmark & \checkmark &\checkmark & \checkmark\\ \hline
      \multirow{5}{*}{Answer Type} & \Block{1-2}{Entity} & ~  & \checkmark & \checkmark & \checkmark & \checkmark &\checkmark &\checkmark & \checkmark \\ \cline{2-10} 
        ~ & \multirow{3}{*}{\shortstack{Time}} & Year & \checkmark & \checkmark & ~ & \checkmark &\checkmark &\checkmark & \checkmark \\ 
        ~ & ~ & Month & \checkmark & \checkmark & ~ & ~ & &\checkmark & \\ 
        ~ & ~ & Day & \checkmark & \checkmark & \checkmark & ~ & &\checkmark & \\  \cline{2-10}
         & \Block{1-2}{Boolean} & & ~ & ~ & ~ & ~ & & ~ & \checkmark \\  \hline
        \multirow{2}{*}{Complexity} & \Block{1-2}{Simple} & &  \checkmark & \checkmark & \checkmark & \checkmark & &\checkmark & \checkmark\\ \cline{2-10}
        ~ &  \Block{1-2}{Complex} & & \checkmark &  \checkmark  &\checkmark & \checkmark & \checkmark &\checkmark & \checkmark\\ 
        \bottomrule
    \end{NiceTabular}}
\end{table*}

\section{Decomposition-based TKG Embedding Methods}
\label{TKGembedding}
Temporal Knowledge Graph Embedding attempts to learn the representations of entities, relations, and timestamps in low-dimensional latent feature spaces while preserving certain properties of the original graph.

Formally speaking, given a TKG $\mathcal{K} = (\mathcal{E,R,T,F})$, TKG embedding methods typically learn $D$-dimential vectors $\mathbf{e}_\epsilon, \mathbf{v}_r, \mathbf{t}_\tau \in \mathbb{R}^D$ for each $\epsilon \in \mathcal{E}$, $r\in\mathcal{R}$ and $\tau\in \mathcal{T}$. These emebdding vectors are learned such that each valid fact $(s,r,o,\tau) \in \mathcal{F}$ is scored higher than an invalid fact $\left(s^{\prime}, r^{\prime}, o^{\prime}, \tau^{\prime}\right) \notin \mathcal{F}$ through a scoring function $\phi(·)$, i.e., $\phi\left(\mathbf{e}_s, \mathbf{v}_r, \mathbf{e}_o, \mathbf{t}_\tau\right)>\phi\left(\mathbf{e}_{s^{\prime}}, \mathbf{v}_{r^{\prime}}, \mathbf{e}_{o^{\prime}}, \mathbf{t}_{\tau^{\prime}}\right)$~\cite{kazemiRepresentationLearningDynamic2020}. Different TKG embedding models have been proposed so far to efficiently learn the representations of TKGs and perform TKG completion as well as inference~\cite {zhangSurveyTemporalKnowledge2024b}.

Decomposition-based methods are one line of the TKG embedding methods. A tensor is a multidimensional array~\cite{kolda2009tensor}.
Tensor decomposition has three applications: dimension reduction, missing data completion, and implicit relation mining, which meet the needs of representation learning.~\cite{caiSurveyTemporalKnowledge2024}

For the temporal knowledge graph, the quadruple can be represented by an order of four tensors, and each tensor dimension is the head entity, relation, tail entity, and timestamp, respectively.

\subsection{TComplex and TNTComplex}
Lacroix et al.~\cite{lacroixTensorDecompositionsTemporal2020} utilise tensor decomposition and proposed the TComplEx and TNTComplEx based on ComplEx~\cite{trouillonComplexEmbeddingsSimple2016}. 

ComplEx is a KGE method. For each fact $(s, r, o)$ in static KG, all entities and relations are embedded into a Complex vector space $\mathbb{C}$. In detail, ComplEx represents facts by three sets of embedding vectors: one for head entities, another for tail entities, and a third for relations. The score function $\phi (s, r, o)$ is defined as: 

$$
\begin{aligned}
\phi(s, r, o) & =\mathfrak{R}\left(<\boldsymbol{e}_s, \boldsymbol{v}_r, \overline{\boldsymbol{e}}_o>\right) \\
& =\mathfrak{R}\left(\sum_{i=1}^d \boldsymbol{e}_{s_i} \boldsymbol{v}_{r_i} \overline{\boldsymbol{e}}_{o_i}\right)
\end{aligned}
$$

where $\bar{\boldsymbol{e}}$ represents the complex conjugate of $\boldsymbol{e}$, $\mathfrak{R}$ denotes adopting the real part of the complex number in the score function.

TComplEx extended ComplEx by increasing the dimensions of the fact tensor. The timestamp of temporal facts is also denoted as complex vector $t_\tau \in \mathbb{C}^{d\times1}$. The score function  $\phi(s, r, o, \tau)$ is transformed as follows:

$$
\begin{array}{r}
\phi(s, r, o, \tau)=\mathfrak{R}\left(<\boldsymbol{e}_s, \boldsymbol{v}_r, \overline{\boldsymbol{e}}_o, \boldsymbol{t}_\tau>\right) \\
=\mathfrak{R}\left(<\boldsymbol{e}_s, \boldsymbol{v}_r \odot \boldsymbol{t}_\tau, \overline{\boldsymbol{e}}_o>\right)
\end{array}
$$
$\odot$ denotes the element-wise product.  

TComplEx employs additional regularizers to improve the quality of the learned embeddings, such as enforcing close timestamps to have similar embeddings (closeness of time). The embedding learning procedure makes TComplEx a suitable method for inferring missing facts such as $(s, r, ?, t)$ and $(s, r, o, ?)$ over an incomplete TKG. Due to its aforementioned benefits, it is used by most TKG Embedding-based TKGQA methods.

Lacroix et al.~\cite{lacroixTensorDecompositionsTemporal2020} also discovered that in the real world, some relationships between entities remain constant, such as ``is located in'', where Shanghai always remains in China. Therefore, TNTComplEx introduces the non-temporal part together with TComplEx. It decomposes a tensor into a sum of tensors with and without time information. This allows the sharing of parameters between parts with and without time information. To represent adjacent timestamps more closely, a penalty is applied to the norm of time embeddings using the smoothness theory.

The score function for TNTComplEx denotes as $\mathfrak{R}\left(<\boldsymbol{e}_s, \boldsymbol{v}_r, \overline{\boldsymbol{e}}_o, \boldsymbol{t}_\tau> + e_s, <\boldsymbol{e}_s, \boldsymbol{u}_r, \overline{\boldsymbol{e}}_o, 1>\right)$, where $ \boldsymbol{u}_r$ represents the temporal agnostic relation representation.

Many other effective TKGE methods may be applied to the TKGQA task~\cite{liuTLogicTemporalLogical2022a,yuanTRHyTETemporalKnowledge2022, xiaMetaTKGLearningEvolutionary2023}, leaving it for the reader to explore.

\end{document}